%% file: main.tex
\documentclass{article}

\input{defs.tex}

\usepackage[utf8]{inputenc} 
\usepackage[T1]{fontenc}    
\usepackage{hyperref}       
\usepackage{url}            
\usepackage{booktabs}       
\usepackage{amsfonts}       
\usepackage{nicefrac}       
\usepackage{microtype}      
\usepackage{xcolor}         

\title{Reinforcement Learning\\ with Non-Exponential Discounting}

\author{%
  Matthias Schultheis\\
  Centre for Cognitive Science\\
  Technische Universität Darmstadt\\
  matthias.schultheis@tu-darmstadt.de \\
  \And
  Constantin A.~Rothkopf \\
  Centre for Cognitive Science\\
  Technische Universität Darmstadt\\
  constantin.rothkopf@tu-darmstadt.de \\
  \And
  Heinz Koeppl \\
  Centre for Cognitive Science\\
  Technische Universität Darmstadt\\
  heinz.koeppl@tu-darmstadt.de \\
}

\begin{document}

\maketitle

\input{sections/0_abstract}
\input{sections/1_introduction}
\input{sections/2_background}
\input{sections/3_method}
\input{sections/4_experiments}
\input{sections/5_conclusion}

\begin{ack}
We thank Bastian Alt for insightful discussions and for providing feedback on the first draft.
This work was supported by the high-performance computer Lichtenberg at the NHR Centers NHR4CES at TU Darmstadt and the project ``Whitebox'' funded by the Priority Program LOEWE of the Hessian Ministry of Higher Education, Science, Research and Art.
\end{ack}

{
\small
\bibliographystyle{unsrtnat} 
\bibliography{references}
}

\input{sections/6_checklist}
\input{sections/7_appendix}

\end{document}

%% file: defs.tex
\PassOptionsToPackage{numbers}{natbib}

\usepackage[final]{neurips_2022}

\usepackage[utf8]{inputenc} 
\usepackage[T1]{fontenc}    
\usepackage[colorlinks=true,linkcolor=blue,citecolor=blue]{hyperref}       
\usepackage{url}            
\usepackage{booktabs}       
\usepackage{amsfonts}       
\usepackage{nicefrac}       
\usepackage{microtype}      
\usepackage{xcolor}         

\usepackage{amsmath}        
\usepackage{bm}             
\usepackage[ruled,vlined]{algorithm2e}
\usepackage[capitalise]{cleveref}

\usepackage[shortlabels,inline]{enumitem}  
\newlist{inlineitemize}{enumerate*}{1}
\setlist[inlineitemize]{label=(\roman*)}

\usepackage{float}
\usepackage{wrapfig}
\usepackage{tikz}           
\usetikzlibrary{bayesnet}   
\usepackage{subcaption}     
\usepackage{layouts}        
\graphicspath{{figures/}{fig/}}

\usepackage{amsthm}
\newtheorem{theorem}{Theorem}
\newtheorem*{theorem-non}{Theorem}

\usepackage{hyphenat} 

\makeatletter
\newcommand{\maketitlenew}{\@maketitle}
\makeatother

\DeclareMathOperator{\tr}{\mathrm{tr}\,}

\DeclareMathOperator*{\argmax}{\arg\max}

\DeclareMathOperator{\R}{\mathbb{R}}

\DeclareMathOperator{\E}{\mathbb{E}\,}

\newcommand{\innermid}{\nonscript\;\delimsize\vert\nonscript\;}
\newcommand{\activatebar}{%
  \begingroup\lccode`\~=`\|
  \lowercase{\endgroup\let~}\innermid 
  \mathcode`|=\string"8000
}

\newcommand{\eqdef}{=\mathrel{\mathop:}}
\newcommand{\defeq}{\mathrel{\mathop:}=}

\newcommand\littleO{o}
\newcommand\given[1][]{\:#1\vert\:}

\DeclareMathOperator{\GamDis}{\mathrm{Gamma}}

\DeclareMathOperator{\UniformDis}{\mathrm{Uniform}}

\newcommand{\td}{\mathrm{d}}
\newcommand{\Sx}{\mathbf{x}}
\newcommand{\SX}{\mathbf{X}}
\newcommand{\Suc}{\mathbf{u}}
\newcommand{\Su}{\mathbf{u}}
\newcommand{\Sxset}{\mathcal{X}}
\newcommand{\Suset}{\mathcal{U}}
\newcommand{\SR}{R}
\newcommand{\Spi}{\mathcal{\pi}}
\newcommand{\Sdt}{\Delta t}
\newcommand{\Sti}{\tau}
\newcommand{\SW}{\mathbf{W}}

\newcommand{\SG}{G}
\newcommand{\Sp}{{\bm{\theta}}}
\newcommand{\Rplus}{\R^+_0}

\newcommand{\Vx}{V_{\Sx}}
\newcommand{\Vt}{V_{t}}
\newcommand{\Vxx}{V_{\Sx \Sx}}
\newcommand{\V}{V}

\usepackage{acronym}
\acrodef{RL}{reinforcement learning}
\acrodef{IRL}{inverse reinforcement learning}
\acrodef{HJB}{Hamilton–Jacobi–Bellman}
\acrodef{SMDP}{semi-Markov Decision Process}
\acrodefplural{SMDP}{semi-Markov Decision Processes}
\acrodef{MDP}{Markov Decision Process}
\acrodefplural{MDP}{Markov Decision Processes}

%% file: sections/0_abstract.tex
\begin{abstract}
Commonly in reinforcement learning (RL), rewards are discounted over time using an exponential function to model time preference, thereby
bounding the expected long-term reward. In contrast, in economics and psychology, it has been shown that humans often adopt a hyperbolic discounting scheme, which is optimal when a specific task termination time distribution is assumed.
In this work, we propose a theory for continuous-time model-based reinforcement learning generalized to arbitrary discount functions. 
This formulation covers the case in which there is a non-exponential random termination time.
We derive a \ac{HJB} equation characterizing the optimal policy and describe how it can be solved using a collocation method, which uses deep learning for function approximation.
Further, we show how the inverse RL problem can be approached, in which one tries to recover properties of the discount function given decision data. We validate the applicability of our proposed approach on two simulated problems. Our approach opens the way for the analysis of human discounting in sequential decision-making tasks. 
\end{abstract}

%% file: sections/1_introduction.tex
\acresetall
\section{Introduction}
An often observed phenomenon in humans and animals is that they prefer rewards rather sooner than later \cite{o2000economics}. It comes with no surprise that an animal searching for food aims to find it as soon as possible and employees working in a company prefer to be paid after each month instead of after each year. In behavioral experiments, it was observed that people are even willing to pay a price to receive rewards earlier \cite{frederick2002time}.
It can be concluded that rewards become of less value in the future, a phenomenon which is known as discounting \cite{frederick2002time}.

Modeling and experimentally inferring discounting functions, which describe how values are discounted over time, 
already have a long history in economics and psychology, where many different functional forms have been proposed \cite{mckerchar2009comparison, strotz1955myopia, thaler1981some}. While from an economic perspective, a fixed interest rate seems reasonable, leading to an exponential discount function, human behavior is oftentimes better described by a hyperbolic curve \cite{mazur1987adjusting}. 
The reason for this is that human decisions are typically not consistent regarding shifting rewards in time, a phenomenon named preference reversal:
We might find that a subject prefers a smaller reward on the same day over a larger reward one day after. However, when given the choice between the same smaller reward in 365 days, compared to the same larger reward in 366 days, the subject is more likely to be willing to wait one day more for the larger reward.

While some literature has branded the observed discounting behavior as not being rational \cite{fisher1930theory}, there has been an increasing amount of work identifying circumstances under which this behavior is indeed optimal \cite{dasgupta2005uncertainty, takahashi2005loss, ray2011positive}. One widely-adopted theory rationalizing hyperbolic discounting is to assume a constant risk for the reward to become unavailable but with the risk being uncertain \cite{sozou1998hyperbolic}. Under this condition, one should adapt the preference over time, as with time the expected risk decreases.

Discounting is also widely applied in the field of reinforcement learning (RL) and optimal control \cite{sutton2018reinforcement, bertsekas2019reinforcement, doya2000reinforcement}.
First, when modeling infinite horizon time objectives, a discount function is needed to make the expected long-term reward objective well-defined, as otherwise, it would become infinite.
Second, for autonomous agents, it also makes sense to model a preference for earlier rewards in order to find shortest paths to save time and energy. Third, the discount function can be interpreted as the probability of termination inducing a specific end-time distribution \cite{puterman1990markov,sutton2018reinforcement}.

Despite these obvious connections, discounting models of psychology and reinforcement learning have remained mostly independent with few exceptions \cite{alexander2010hyperbolically, fedus2019hyperbolic}. Generalizing reinforcement learning to a broader range of discount functions would enable solutions for applications in which the end time follows a specific distribution.
On the other hand, methods for determining optimal decisions in sequential decision-making tasks with general end-time distributions would provide tools which can help to explain human decision-making under uncertainty.

In this work, we present a theory for model-based reinforcement learning in continuous time based on non-exponential discount functions. First, we investigate the conditions under which the objective of maximizing the long-term reward formulated with hyperbolic discounting is well-defined. Second, we derive a \ac{HJB}-type equation for a general discount function and describe how to solve it to obtain the optimal policy.
Third, we provide an approach to tackle the inverse reinforcement learning (IRL) problem, to estimate parameters of the discount function given decision data.
Finally, we show the applicability of our proposed method on two simulated problems.

\section{Related work}
Optimal control in continuous time and space has a long history with many classical works \cite{stratonovich1968conditional, kushner2001numerical, fleming2006controlled, pontryagin1987mathematical}. Continuous-time reinforcement learning formulations have been developed \cite{doya2000reinforcement} and various solution methods have been proposed \cite{vamvoudakis2010online, baird1994reinforcement}. Solution approaches that solve the \ac{HJB} equation directly include linearization techniques \cite{jacobson1968new, tassa2007receding}, path integral formulations \cite{kappen2005path, theodorou2010generalized, sutton2018reinforcement}, and collocation-based methods \cite{simpkins2009practical}. In recent years, it has become increasingly popular to use neural networks for function approximation to solve the \ac{HJB} equation \cite{tassa2007least, sirignano2018dgm, lutter2020hjb, han2018solving, alt2020pomdps}.

Non-exponential discounting was considered in literature in behavioral economics, psychology, and neuroscience \cite{strotz1955myopia, thaler1981economic, frederick2002time}. For humans and animals, preference reversal behavior has been described \cite{ainslie1981preference, green1994temporal, thaler1981some} and different functional forms for the discount function have been proposed \cite{mckerchar2009comparison, andersen2014discounting}. 
In most works, only non-sequential decision-making tasks have been studied, i.e., situations with a single decision for each independent trial. In this line of work, the method of limits has been used to elicit discount functions based on binary decisions \cite{richards1997determination, coller1999eliciting, andersen2008eliciting} or an adjustment method has been applied \cite{noor2009hyperbolic}.
Some methods have been proposed that specifically aim to efficiently estimate the discount function for binary choice responses \cite{cavagnaro2016functional, chang2019modeling}. 
There have been a few exceptions for which decision trials were actually dependent but could still be considered independently for the analysis \cite{schweighofer2006humans, seinstra2018rate}.
Further, there has been research aiming to find rational explanations for the encountered discounting behavior \cite{dasgupta2005uncertainty, takahashi2005loss, ray2011positive}, e.g., by assuming uncertainty about a constant hazard \cite{sozou1998hyperbolic}.

Decision processes with non-exponential time distributions have been studied in the field of \acp{SMDP} \cite{howard1963semi, korolyuk1975semi, ross1970average, bradtke1994reinforcement}, where transition times between discrete states can follow arbitrary distributions. MDPs with quasi-hyperbolic discounting, for which all future rewards are additionally discounted by a constant factor, have been addressed in \cite{jaskiewicz2021markov, bjork2010general, bjork2014theory}. They can be used to model preference reversal but are limited to the specific form assumed. Stochastic processes with quasi-hyperbolic discounting have been considered in financial economics for portfolio management \cite{grenadier2007investment, zou2014finite, chunxiang2016optimal, chen2018optimal}.

More closely related to our work, \citet{fedus2019hyperbolic} presented an approximate method for solving MDPs with hyperbolic discounting. They approached the problem by solving the corresponding exponentially-discounted problem for many different discount factors and combining the results. Part of the approximation, however, is that the value function and policy are static, failing to model preference reversals over time. Another line of work considers MDPs with discount factors that are coupled to the value function to imitate hyperbolic discounting behavior \cite{alexander2010hyperbolically, van2019general}. As time is only considered indirectly through the magnitude of the value function, these approaches cannot be used for finding the solution to a given specific discount function or eliciting the discount function from data.

Finally, inverse reinforcement learning approaches have been mainly used to learn reward functions given data \cite{ng2000algorithms, abbeel2004apprenticeship, ziebart2008maximum, garg2021iq}. These methods were also applied to learn properties of human behavior \cite{mombaur2010human, rothkopf2013modular,muelling2014learning, schultheis2021inverse}. Other inverse approaches have focused on learning dynamics models \cite{golub2013learning} or learning rules \cite{ashwood2020inferring} from sequential decision-making data.

%% file: sections/2_background.tex
\section{Background}

\subsection{Survival analysis}
\label{sec:survival_analysis}
In survival analysis \cite{aalen2008survival}, one is interested in the duration until events occur. In its classical form one considers a single event, for which the duration can be described as a continuous random variable $T$ with cumulative distribution function $F(t) = P(T \leq t)$ and probability density function $f(t)$, where $t \in \Rplus$ denotes the elapsed time. In survival analysis literature, $F(t)$ is known as the failure function and the survival function is defined as $S(t) = 1 - F(t) = P(T > t)$. 
The survival function is monotonically decreasing and has the properties $S(0) = 1$ and $\lim_{t \to \infty} S(t) = 0$.
By the conditioning rule, one finds $P(T > t_1 \given T > t_0) = S(t_1)/S(t_0)$ if $t_1 > t_0$.
The hazard rate is defined as $\alpha(t) = \lim _{\Sdt \rightarrow 0} \frac{1}{\Sdt} P(t \leq T < t+\Sdt \given t \leq T)$, yielding the relations
\begin{align}
\label{eq:surv_analysis}
    \alpha(t)=\lim _{\Sdt \rightarrow 0} \frac{1}{\Sdt} \frac{S(t)-S(t+\Sdt)}{S(t)}= -\frac{S^{\prime}(t)}{S(t)} \quad\text{and} \quad S(t) = \exp\left(-\int_{0}^t \alpha(\Sti) \, \td \Sti\right).
\end{align}
For a constant hazard rate $\alpha(t) = \lambda$, one finds $S(t) = \exp(-\lambda t)$, which can be shown to be the unique memory-less survival function \cite{aalen2008survival}, i.e., $P(T > t + \Sdt \given T>t) = P(T > \Sdt)$ for all $t, \Sdt \in \Rplus$.

\subsection{Discounting and preference reversal}
\label{sec:backgorund_discounting}
We consider the setting in which a subject collects a single reward and shows a form of time preference, i.e., rewards are desired rather sooner than later. We model the value of a reward $r \in \R$ as a function $L: \R \times \Rplus \to \R$ with $L(r, t) = S(t) \cdot r$, where $S(t)$ is the discount function decreasing with time $t \in \Rplus$.
With the convention that $S(t_0) = 1$ and $\lim_{t \to \infty} S(t) = 0$, we can regard $S(t)$ as the survival function (\cref{sec:survival_analysis}), with the interpretation that the reward becomes unavailable with hazard rate $\alpha(t)$ \cite{sozou1998hyperbolic}.

When assuming a constant hazard rate $\alpha(t) = \lambda$, we say that the subject discounts exponentially, as $S(t) = \exp(-\lambda t)$. By the memory-less property, we have that if the subject prefers reward $r_1$ after $t_1$ over $r_2$ after $t_2$, she or he would remain consistent with the election if we presented the choice again later in time.
On the other hand, if we assume a constant but unknown hazard rate $\lambda$ with belief $p(\lambda) = \GamDis(\lambda ; \alpha_0, \beta_0)$, we obtain a hyperbolic form for the expected survival function:
\begin{align}\label{eq:hyp_sf}
    S(t; \alpha_0, \beta_0) &= \int_{\lambda} \exp (-\lambda t) \, p(\lambda) \, \td \lambda
    =\frac{1}{(\frac{t}{\beta_0} + 1)^{\alpha_0}}
\end{align}
The posterior belief over $\lambda$ at a later point in time can be derived using Bayes rule and is given by $p(\lambda \given t) = \GamDis(\lambda; \alpha_0, \beta_0 + t)$. 
The expected hazard rate $\alpha(t)$ is given by the posterior mean,
\begin{align}\label{eq:hyp_hazard}
    \alpha(t) = \int_{\lambda} \lambda \, p(\lambda \given t) \, \td \lambda
    =\frac{\alpha_0}{\beta_0 + t}.
\end{align}
For discount functions other than the exponential, such as the hyperbolic discount function in \cref{eq:hyp_sf}, the hazard rate varies over time and preferences among options may change.

\subsection{Optimal control}
\label{sec:opt_control}
In stochastic optimal control \cite{hanson2007applied}, we consider a Markovian system with continuous state $\Sx (t) \in \R^{n}$ evolving according to the stochastic differential equation (SDE) $ \td {\SX}(t) = f(\SX(t), \Suc(t)) \, \td t + \SG (\SX(t), \Suc(t)) \, \td \SW(t)$ where $t \in \Rplus$ denotes time, $f: \Sxset \times \Suset \to \Sxset$ the drift function, $\SG: \Sxset \times \Suset \to \Sxset \times \R^m$ the dispersion matrix, and $\SW(t) \in \R^m$ $m$-dimensional Brownian motion. The goal of optimal control is to determine the control inputs $\Suc(t) \in \Suset$ given the current state $\Sx$ at time $t$, in order to maximize the expected long term discounted reward with reward function $\SR: \Sxset \times \Suset \to \R$. The solution is characterized by the optimal value function,
\begin{align}
    \label{eq:oc_value_fun}
    V^*(\Sx) = \max_{\Suc_{[t, \infty)}} \E \left[\int_t^\infty \exp({-\lambda(\tau - t)}) \, \SR(\SX(\tau), \Suc(\tau)) \, \mathrm{d}\tau \given[\Big] \SX(t) = \Sx\right],
\end{align}
where maximization is carried out over all trajectories $\Suc_{[t, \infty)} \defeq \{\Su(\Sti) \}_{\Sti \in [t, \infty]}$. The quantity $\exp(-\lambda t)$ is the discount factor, which ensures convergence of the integral and models a preference for earlier rewards. $\lambda$ can be interpreted in terms of survival analysis as the hazard rate for termination (cf.\ \cref{sec:survival_analysis}). 
According to the principle of optimality, the stochastic Hamilton-Jacobi-Bellman (HJB) equation, given by
\begin{align*}
    \lambda V^*(\Sx) = \max_{\Suc \in \Suset} \left[ \SR(\Sx, u) + \Vx^*(\Sx)^T f(\Sx, \Su) + \frac{1}{2} \tr{\left\{\Vxx^*(\Sx) \SG(\Sx, \Suc) \SG(\Sx, \Suc)^T \right\}} \right],
\end{align*}
provides a condition for the optimal value function. Here, partial derivatives are denoted by the index notation, i.e., $\Vx$ denotes the partial derivative of $V$ w.r.t.\ $\Sx$ and $\Vxx$ the respective Hessian. Note that the optimal value function depends on time only indirectly through the state $\Sx$ due to the Markov property and the memory-less discount function. This dependence also applies to the optimal policy $\Spi^*: \Sxset \to \Suset$, given by the maximizer of the right-hand side of the HJB equation, i.e.,
\begin{align*}
    \pi(\Sx) = \argmax_{\Suc \in \Suset} \left[ \SR(\Sx, u) + V^*_{\Sx}(\Sx)^T f(\Sx, \Suc) + \frac{1}{2}V^*_{\Sx \Sx}(\Sx) \SG(\Sx, \Suc) \SG(\Sx, \Suc)^T \right].
\end{align*}

%% file: sections/3_method.tex
\section{Reinforcement learning with general discount function}
\label{sec:rl_nonexp}

We consider a system as in stochastic optimal control (cf.\ \cref{sec:opt_control}) with continuous state space $\R^{n}$ and finite set of controls $\Suset$.
Instead of an exponential discount function, we allow for general survival functions $S(t)$ based on a time-dependent hazard rate $\alpha(t)$. This setting generalizes the discounting formulation of \cref{sec:backgorund_discounting} to sequential decisions.
As we will see later, the resulting value function will become time-dependent in contrast to \cref{sec:opt_control} and we can further allow for time-dependent dynamics and reward without increasing complexity. We therefore assume for the state evolution the SDE $ \td {\SX}(t) = f(\SX(t), \Suc(t), t) \, \td t + \SG (\SX(t), \Suc(t), t) \, \td \SW(t)$ with $f: \Sxset \times \Suset \times \Rplus \to \Sxset$ and $\SG: \Sxset \times \Suset\times \Rplus \to \Sxset \times \R^m$.
The objective function measuring the total expected discounted reward is given by
\begin{align}
\label{eq:method_objective_fun}
    J\left(\Su_{[0, \infty)}\right) = \E \left[ \int_0^\infty S(\Sti) \, \SR(\SX(\Sti), \Su(\Sti), \Sti) \, \mathrm{d}\Sti  \right],
\end{align}
with time-dependent reward function $\SR: \Sxset \times \Suset \times \Rplus \to \R$.
Analogous to \cref{sec:opt_control}, we define the expected reward-to-go as the value function
\begin{align}
\label{eq:method_v_fun}
    V^*(\Sx, t) = \max_{\Su_{[t, \infty)}} \E \left[ \int_t^\infty \frac{S(\Sti)}{S(t)} \SR(\SX(\Sti), \Su(\Sti), \Sti) \, \mathrm{d}\Sti \given[\Big] \SX(t) = \Sx \right],
\end{align}
where $S(\Sti)/S(t)$ is the probability of survival until time $\Sti$, conditioned on the fact that one already has survived until time $t$ (cf.\ \cref{sec:survival_analysis}).
In contrast to \cref{eq:oc_value_fun}, the value function becomes time-dependent through the general survival function and also the optimal policy depends on time.

\subsection{Technical requirements of model and discount function}
First, the stochastic process defined in \cref{sec:rl_nonexp} must have a strongly unique solution, which is the case if $f$ and $G$ grow at most linearly in $\SX$ and are Lipschitz continuous in the same variable \cite{sarkka2019applied}. Further, the optimal value function is only in rare cases smooth enough to be a solution in the ``classical'' sense. Instead, one considers a viscosity solution \cite{fleming2006controlled}, which satisfies the HJB equation in an appropriate generalized sense. A sufficient condition for the existence of such a solution is that $f$ and $G$ are continuous with bounded continuous first derivatives w.r.t. $\SX$ and $t$, and bounded. $\SR$ and $S$ need to be continuous and grow at most polynomially in $\SX$ and $t$ in an absolute sense. More details on the existence and uniqueness of viscosity solutions can be found in \cite{fleming2006controlled}. 
A third requirement is that the integral in the definition of the value function \cref{eq:method_objective_fun} converges. For the case of a hyperbolic discount function as in \cref{eq:hyp_sf}, we find the following theorem:
\begin{theorem}
Consider the hyperbolic discount function in \cref{eq:hyp_sf}. If the reward function $\SR(\Sx, \Su, t)$ is bounded above for all $(\Sx, \Su, t) \in \Sxset \times \Suset \times \Rplus$, and $\alpha_0 > 1$, the value function defined in equation \cref{eq:method_v_fun} is well-defined. If $\SR(\Sx, \Su, t)$ is bounded below for all $(\Sx, \Su, t) \in \Sxset \times \Suset\times \Rplus$, and $\alpha_0 \leq 1$, the value function is not well-defined as it becomes infinite.
\end{theorem}
\renewcommand\qedsymbol{}
\begin{proof}
See \cref{app:convergence}.
\end{proof}
In the examples considered later, we will assume a bounded reward function and a hyperbolic discount function with $\alpha_0 > 1$, for which \cref{eq:method_objective_fun} and \cref{eq:method_v_fun} are well-defined.

\subsection{HJB equation for a general discount function}
\label{sec:hjb_general}
In the following, we give a brief overview of the derivation of the HJB equation for a general discount function. A more detailed derivation is provided in \cref{app:derivation_hjb}.
First, we split the integral in \cref{eq:method_v_fun} into two terms such that we obtain a recursive formulation of the value function:
\begin{align*}
    V^*(\Sx, t) = \max_{\Su_{[t, t+ \Delta t]}} \E &\left[ \int_t^{t+\Delta t} \frac{S(\Sti)}{S(t)} \SR(\SX(\Sti), \Su(\Sti), \Sti) \, \td\Sti \right.\\
    &+\Biggl. \frac{S(t + \Delta t)}{S(t)} V^*(\SX(t+ \Delta t), t + \Delta t) \given[\Big] \SX(t) = \Sx \Biggr]
\end{align*}
For the second term in the expectation, we apply a Taylor expansion and It\^{o}'s formula \cite{hanson2007applied} and obtain
\begin{align*}
    &V^*(\SX(t+ \Sdt), t + \Sdt) = V^*(\SX(t), t) + \int_t^{t+\Sdt} \Vx^*(\SX(\Sti), \Sti) \, f(\SX(\Sti), \Su(\Sti), \Sti)\, \td\Sti \\
    &\qquad+ \int_t^{t+\Sdt} \Vx^*(\SX(\Sti), \Sti) \, \SG(\SX(\Sti), \Su(\Sti), \Sti) \, \td\SW(\Sti) + \int_t^{t+\Sdt} \Vt^*(\SX(\Sti), \Sti) \, \td\Sti \\
    &\qquad+ \int_t^{t+\Sdt} \frac{1}{2} \tr{\left\{\Vxx^*(\SX(\Sti), \Sti) \, \SG(\SX(\Sti), \Su(\Sti), \Sti) \, \SG(\SX(\Sti), \Su(\Sti), \Sti)^T \right\}} \, \td\Sti + \littleO(\Sdt).
\end{align*}
Plugging this result into the equation above, dividing both sides by $\Sdt$, and taking the limit $\Sdt \to 0$, as well as calculating the expectation w.r.t.\ $\SW(t)$ leads to 
the desired HJB equation
\begin{align}
\label{eq:hjb}
\begin{split}
    \alpha(t) V^*(\Sx, t) = \max_{\Su} &\biggl[ \SR(\Sx, \Su, t)+ \Vt^*(\Sx, t) + \Vx^*(\Sx, t) \, f(\Sx, \Su, t) \biggr. \\
    & \quad +  \left.\frac{1}{2} \tr{\left\{\Vxx^*(\Sx, t) \, \SG(\Sx, \Su, t) \, \SG(\Sx, \Su, t)^T \right\}} \right],
\end{split}
\end{align}
where $\alpha(t)$ can be recognized to be the hazard rate corresponding to the survival function $S(t)$.
We define the r.h.s.\ of the HJB equation in \cref{eq:hjb} without the maximization w.r.t.\ the action as
\begin{align*}
\begin{split}
    Q(\Sx, \Su, t) \defeq \ &\SR(\Sx, \Su, t)+ \Vt^*(\Sx, t) + \Vx^*(\Sx, t) \, f(\Sx, \Su, t) \\
    &+ \frac{1}{2} \tr{\left\{\Vxx^*(\Sx, t) \, \SG(\Sx, \Su, t) \, \SG(\Sx, \Su, t)^T \right\}},
\end{split}
\end{align*}
so that the optimal policy is given by $\pi^*(\Sx, t) = \argmax_\Su Q(\Sx, \Su, t)$. Later on, we will consider hyperbolic discounting, for which the hazard rate $\alpha(t)$ is given by \cref{eq:hyp_hazard}, i.e., $\alpha(t) = \alpha_0 / (\beta_0 + t)$.

\subsection{Solving the HJB equation}
\label{sec:collocation}
In order to solve the HJB equation in \cref{eq:hjb}, which is a PDE, we apply a collocation-based method \cite{simpkins2009practical, tassa2007least, lutter2020hjb}. To do so, we first reformulate the HJB equation as
\begin{align}
\label{eq:hjb_constraint}
\begin{split}
    E(\V, \Sx, t) \defeq - \alpha(t) \V(\Sx, t) + \max_{\Su} &\biggl[\SR(\Sx, \Su, t)+ \Vt(\Sx, t) + \Vx(\Sx, t) \, f(\Sx, \Su, t) \biggr.\\
    &+ \left. \frac{1}{2} \tr{\left\{\Vxx(\Sx, t) \,\SG(\Sx, \Su, t) \,\SG(\Sx, \Su, t)^T \right\}} \right] = 0,
\end{split}
\end{align}
and use a function approximator $V^{\psi}(\Sx, t)$ for $V^*(\Sx, t)$. The parameters $\psi$ of the approximator can be determined by sampling random states $\hat{\Sx}_i$ and time points $\hat{t}_i$ and minimizing $\sum_i E(\V^\psi, \hat{\Sx}_i, \hat{t}_i) ^ 2$ w.r.t.\ $\psi$. Calculating the derivatives $V_\Sx^\psi(\Sx, t)$, $V_t^\psi(\Sx, t)$, $V_{\Sx \Sx}^\psi(\Sx, t)$ and differentiating the objective function w.r.t.\ $\psi$ is straightforward via automatic differentiation when choosing a neural network as function approximator \cite{paszke2019pytorch}.
As $t$ is not bounded, we need to choose a reparametrization of $t$ which maps all $t$ to the interval $[0, 1)$ before feeding the values into the network. More details about the implemented parametrization and application of the collocation method are provided in \cref{app:app_collocation}. The complete algorithm to learn the value function and policy is provided in \cref{alg:collocation}.

\begin{algorithm}[t]
\SetAlgoLined
\KwResult{Optimal value function $V^{\psi}(\Sx, t)$, optimal policy $\pi^{\psi}(\Sx, t)$}
\KwIn{Parameters $\Sp$ of the discount function, system model, number of iterations $K$}
Initialize parameters $\psi$ of neural network for modeling $V^{\psi}(\Sx, t)$\;
\For{$k\leftarrow 0$ \KwTo $K-1$}{
    Sample a batch of states and time points $(\hat{\Sx}, \hat{t})_{i=\{1, \dots, N\}}$\;
    Push $(\hat{\Sx}, \hat{t})_{i=\{1, \dots, N\}}$ through the network to obtain $\hat{V}^{\psi}_{i=\{1, \dots, N\}}$\;
    Use back-propagation to compute $\hat{V}_{\Sx i}^\psi$, $\hat{V}_{t i}^\psi$, $\hat{V}_{\Sx \Sx i}^\psi$\;
    Evaluate $E(\hat{V}_i^\psi, \hat{\Sx}_i, \hat{t}_i)$ in \cref{eq:hjb_constraint} and determine maximizing action $\hat{\Su}_i$ for $i=1, \dots, N$\;
    Use back-propagation to compute the gradient of $\sum_i E(\hat{\V}_i^\psi, \hat{\Sx}_i, \hat{t}_i) ^ 2$ w.r.t.\ $\psi$ \;
    Update $\psi$ using the gradient \;
 }
\textbf{return} $V^{\psi}(\Sx, t)$, $\pi^{\psi}(\Sx, t) = \argmax_\Su Q(\Sx, \Su, t)$
\caption{Computation of the optimal value function and policy for non-exp.\ discounting}\label{alg:collocation}
\end{algorithm}

\subsection{Inverse reinforcement learning for inferring the discount function}
When analyzing human behavior, one might be interested in learning the underlying discount function that led to a set of observed choices. In contrast to standard inverse reinforcement learning settings, where the goal is to learn the underlying reward function $\SR$, here we assume that the reward function is given and the discount function $S(t)$ is unknown and needs to be inferred. To learn the function, we assume a parametric form with parameters $\Sp$.

As given data, we assume the states and time points at which the subject switches from one action to another one, i.e., $\mathcal{D} = \left\{ \Sx, \Su^-, \Su^+, t\right\}_{i=1\dots N}$,
describing that in state $\Sx$ at time $t$ the action $u^-$ is switched to $u^+$. The observed decision maker is assumed to use the optimal policy $\pi(\Sx, t) = \argmax_\Su Q(\Sx, \Su, t)$. 
Shortly before and after switching time $t$, we have $Q(\Sx(t - \Sdt), \Su^-, t -\Sdt) > Q(\Sx(t - \Sdt), \Su^+, t -\Sdt)$ and $Q(\Sx(t + \Sdt), \Su^-, t +\Sdt) < Q(\Sx(t + \Sdt), \Su^+, t +\Sdt)$, respectively, indicating that before $t$ action $\Su^-$ is preferred over $\Su^+$ and vice versa afterwards.
By letting $\Sdt \to 0$, one finds $Q(\Sx(t), \Su^-, t) = Q(\Sx(t), \Su^+, t)$. A sensible objective for the inverse problem is therefore to minimize $\sum_i F(\Sx_i, \Su^-_i, \Su^+_i, t_i)$, defined as the squared difference of both Q values, i.e.,
\begin{align}
    F(\Sx, \Su^-, \Su^+, t) &= \left[ Q(\Sx, \Su^-, t) - Q(\Sx, \Su^+, t) \right]^2\notag\\
    &= \biggl[ \SR(\Sx, \Su^+, t) - \SR(\Sx, \Su^-, t) + V_\Sx(\Sx, t) \left(f(\Sx, \Su^+, t) - f(\Sx, \Su^-, t)\right) \biggr.\label{eq:obj_fun_inverse}\\
    &+ \left.\frac{1}{2} \tr{\left\{V_{\Sx \Sx}(\Sx, t) \left(\SG(\Sx, \Su^+, t) \, \SG(\Sx, \Su^+, t)^T - \SG(\Sx, \Su^-, t) \, \SG(\Sx, \Su^-, t)^T \right)\right\}} \right]^2,\notag
\end{align}
w.r.t.\ the parameters $\Sp$ of the discount function. The optimization needs to take \cref{eq:hjb_constraint} as a constraint into account to ensure that the HJB equation is fulfilled. Note that \cref{eq:obj_fun_inverse} depends indirectly on the parameters $\Sp$ through the definition of the value function. However, the objective function $F$ could also directly depend on $\Sp$, we consider the general case in the following. In principle, the state at which the switching occurs also depends on $\Sp$. Nevertheless, we will assume that the parameters have only a minor influence on the states of switching, so that terms including the variation of $\Sx$ w.r.t.\ $\Sp$ can be neglected.

\subsubsection*{Gradient computation}
For the minimization, it is useful to determine the gradient of the objective function $F$ w.r.t.\ $\Sp$, which can be formulated using the chain rule for total derivatives as
\begin{align}
\label{eq:gradient}
\frac{\td F}{\td \Sp} &= F_\Sp + F_V V_\Sp + F_{V_\Sx}\left( V_\Sx\right)_\Sp + F_{V_{\Sx \Sx}}\left( V_{\Sx \Sx}\right)_\Sp \nonumber\\
&= F_\Sp + F_V V_\Sp + F_{V_\Sx}\left( V_\Sp\right)_\Sx + F_{V_{\Sx \Sx}}\left( V_\Sp\right)_{\Sx \Sx}.
\end{align}
By switching the order of the partial derivatives in the last step, we have obtained an expression for the gradient depending on $V_\Sp$. The partial derivative $V_\Sp$ measures the influence of the discounting parameters on the value function and is not straightforward to evaluate. We can get a PDE for this quantity by observing that the constraint $E(V^*, \Sx, t)$ in \cref{eq:hjb_constraint} is zero everywhere for the optimal value function $V^*$ and therefore its derivative also needs to be zero. This method is known as the forward sensitivity method \cite{rabitz1983sensitivity} and provides an additional PDE for the desired gradient $V_\Sp$:
\begin{align}
\label{eq:pde_param}
0 = \frac{\td E}{\td \Sp} &= E_\Sp + E_V V_\Sp + E_{V_t} \left(V_t \right)_\Sp + E_{V_\Sx} \left(V_\Sx \right)_\Sp + E_{V_{\Sx \Sx}} \left(V_{\Sx \Sx} \right)_\Sp \nonumber\\
&= E_\Sp + E_V V_\Sp + E_{V_t} \left(V_\Sp \right)_t + E_{V_\Sx} \left(V_\Sp \right)_\Sx + E_{V_{\Sx \Sx}} \left(V_\Sp \right)_{\Sx \Sx} \eqdef H(V, V_\Sp, \Sx, t)
\end{align}
To solve the PDE and obtain $V_\Sp$, we can use the same procedure as in \cref{sec:collocation}. 
To obtain the gradient, first one has to solve the HJB equation \cref{eq:hjb} to obtain $V^\psi$, then \cref{eq:pde_param} to determine $V^{\phi}_{\Sp}$. Afterwards, \cref{eq:gradient} gives the gradient to update the parameters $\Sp$ of the discount function. The quantities $V_\Sp$, $\left(V_\Sp \right)_\Sx$, $\left(V_\Sp \right)_{\Sx \Sx}$ for the derivative be computed via automatic differentiation as in \cref{sec:collocation}. The complete algorithm for computing the gradient is presented in \cref{alg:gradient_comp}.

\begin{algorithm}[t]
\SetAlgoLined
\KwResult{Gradient of $F$ (\cref{eq:obj_fun_inverse}) w.r.t.\ $\Sp$ \cref{eq:gradient}}
\KwIn{Parameters $\Sp$ of the discount function, system model, number of iterations $K$}
Determine $V^{\psi}(\Sx, \Su, t)$ using \cref{alg:collocation} \;
Initialize parameters $\phi$ of neural network for modeling $V_{\Sp}^{\phi}(\Sx, t)$\;
\For{$k\leftarrow 0$ \KwTo $K-1$}{
    Sample a batch of states and time points $(\hat{\Sx}, \hat{t})_{i=\{1, \dots, N\}}$\;
    Push $(\hat{\Sx}, \hat{t})_{i=\{1, \dots, N\}}$ through the networks to obtain $\hat{V}^{\phi}_{\Sp i=\{1, \dots, N\}}$ and $\hat{V}^{\psi}_{i=\{1, \dots, N\}}$\;
    Use back-propagation to compute $( \hat{V}^\psi_{\Sp} )_{\Sx i}$, $( \hat{V}^\psi_{\Sp} )_{t i}$, $( \hat{V}^\psi_{\Sp} )_{\Sx \Sx i}$\;
    Evaluate $\sum_i H(\hat{V}^{\psi}_{i}, \hat{V}^{\phi}_{\Sp i}, \hat{\Sx}_i, \hat{t}_i) ^ 2$ in \cref{eq:pde_param} for $i=1, \dots, N$\;
    Use back-propagation to compute the gradient of $\sum_i H(\hat{V}^{\psi}_{i}, \hat{V}^{\phi}_{\Sp i}, \hat{\Sx}_i, \hat{t}_i) ^ 2$ w.r.t.\ $\phi$ \;
    Update $\phi$ using the gradient \;
 }
\textbf{return} $\td F /\td \Sp$ by evaluating \cref{eq:gradient} using $V_{\Sp}^{\phi}(\Sx, t)$
\caption{Computation of the gradient of $F$ w.r.t.\ $\Sp$ for inferring the discount function}\label{alg:gradient_comp}
\end{algorithm}

%% file: sections/4_experiments.tex
\begin{figure}[t!]
    \centering
    \includegraphics[width=\textwidth]{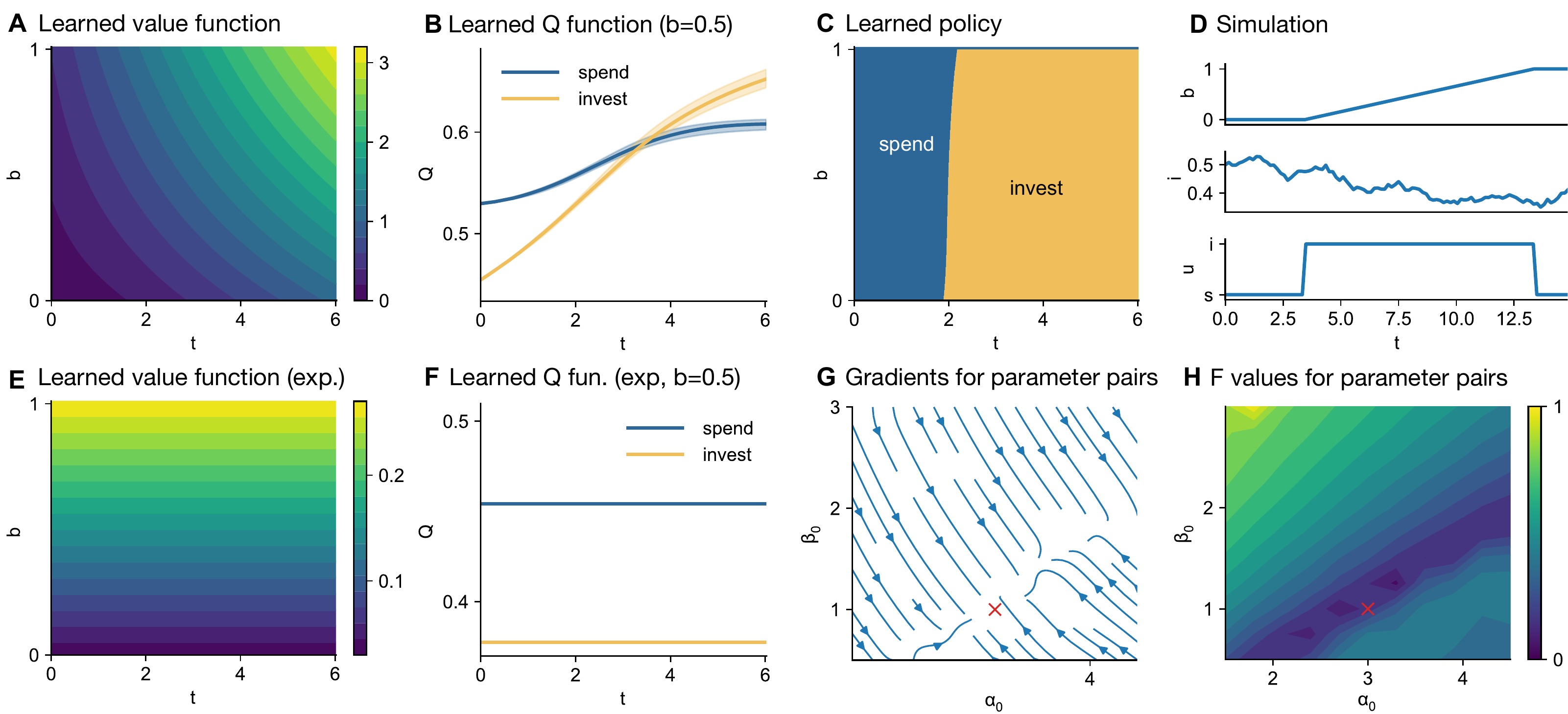}
    \caption{\textbf{Results for the investment problem.} \textbf{A} Learned value function for account balances and time points for hyperbolic discounting. \textbf{B} Learned Q function for both actions over time (median and quantiles for 10 runs). \textbf{C} Learned policy showing preference reversals for all account balances except for $b=1$. \textbf{D} Simulation with account balance (top), interest rate (middle), and action (bottom) over time. \textbf{E} Learned value function for an exponential discount function, which is constant over time. \textbf{G} Learned Q function for an exponential discount function. \textbf{G} Gradients obtained for parameter pairs $[\alpha_0, \beta_0]$ of the discount function given simulated data on a $11\times11$ grid. The parameters used to generate the simulated trajectories are indicated by a red cross. \textbf{H} Values of the objective function $F$ (\cref{eq:obj_fun_inverse}) for parameter pairs on a $11\times11$ grid when applying the IRL method.}
    \label{fig:invest}
\end{figure}
\section{Experiments}
\label{sec:experiments}
We tested our derived method on two simulated problems with a random termination time following a hyperbolic survival function. First, we solved the HJB equation in \cref{eq:hjb} using the collocation method (\cref{alg:collocation}) and computed the optimal policy. Then, for applying the inverse reinforcement learning method, we evaluated the objective function in \cref{eq:obj_fun_inverse} and computed gradients for different parameter sets $\Sp = [\alpha_0, \beta_0]$ of the hyperbolic discount function in \cref{eq:hyp_sf} using \cref{alg:gradient_comp}. For generating data, we randomly sampled starting states and determined the time points at which subjects would switch their action. Afterward, the determined time points were distorted by Gaussian noise.

All proposed methods were implemented in Python using the PyTorch framework \cite{paszke2019pytorch} and are available online\footnote{\url{https://git.rwth-aachen.de/bcs/nonexp-rl}}.
The used hyperparameters are listed in \cref{app:hyperparams}. As tasks, we considered an investment problem and a problem controlling a point on a line. In the following we provide a brief overview of the considered problems, more details can be found in \cref{app:experiments}.

\subsection{Experimental tasks}

In the investment problem, we model a subject having to decide whether to invest her or his income in the bank account leading to future interests (rewards) or to spend the money for immediate reward. We model the state as the current balance of the bank account as well as the current interest rate. When the money is spent (action $\textit{spend}$), the subject receives rewards with a rate of 0.1 but the balance of the account remains unchanged. When the income is invested (action $\textit{invest}$), the balance of the bank account increases with a rate of 0.1 but there is no additional reward. In both cases the subject receives rewards through interests, being proportional to the current balance on the bank account. We assume that the interest rate varies over time following a Gaussian diffusion model. To keep the state bounded, we model a maximum balance for the account.

In the line problem, the task is to control a point along a line. The state consists of the current position of the point. Possible actions are \textit{left} and \textit{right}, which move the point to the respective direction, as well as \textit{stay}, which does not move the point at all. When moving the point, there is a Gaussian diffusion on the position and one has to pay a small action cost (negative reward). We model a state-dependent reward modeling a high reward for distant states on one side and a low reward for close states on the other side. This task can be seen as a continuous extension of the classical discounting tasks. A more detailed description of the functional form is provided in \cref{app:experiments}. 

\begin{figure}[t!]
    \centering
    \includegraphics[width=\textwidth]{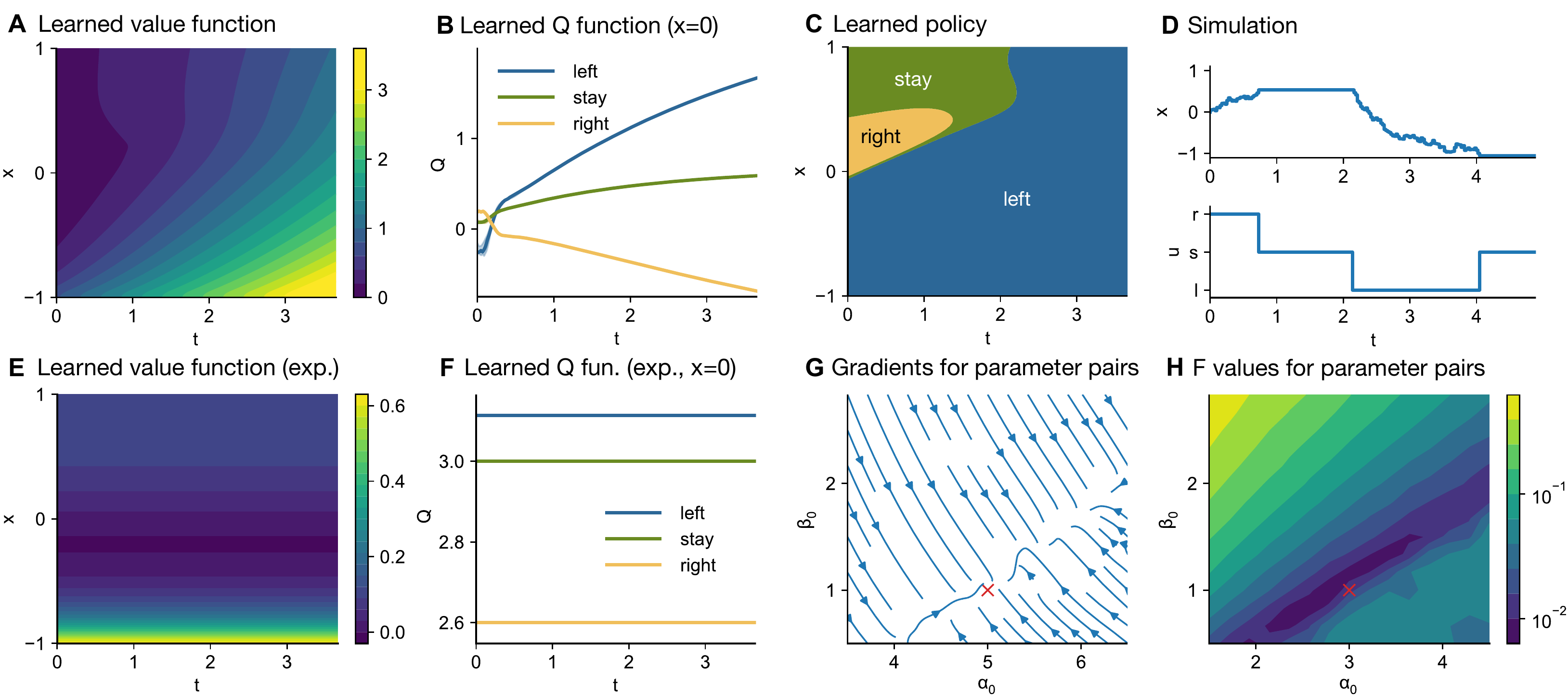}
    \caption{\textbf{Results for the line problem.} \textbf{A} Learned value function for hyperbolic discount function. \textbf{B} Learned Q function for each action over time (median and quantiles for 10 runs) \textbf{C} Learned policy showing preference reversals. \textbf{D} Simulation showing the state (top) and action (bottom) over time. \textbf{E} Learned value function for an exponential discount function. \textbf{G} Learned Q function for an exponential discount function. \textbf{G} Gradients obtained for parameter pairs $[\alpha_0, \beta_0]$ on a $15\times15$ grid given simulated data. The parameter used to generate the simulated trajectories is indicated by a red cross. \textbf{H} Values of the objective function of the IRL methods for parameter pairs on a $15\times15$ grid.}
    \label{fig:line}
\end{figure}

\subsection{Results}
\label{subsec:exp_results}

The proposed method in \cref{alg:collocation} produces plausible value functions for the considered problems. \Cref{fig:invest} A shows the learned value function for the investment problem. The values are increasing with account balance (b), as expected rewards through interests become higher. Further in time, the value also increases, as the hazard rate can be assumed to be lower and one is expected to collect rewards for a longer time. \Cref{fig:invest} B and C shows the learned Q function and policy, respectively. While it is advantageous to spend the income in the beginning, preference reversal occurs when the risk of termination is assumed to be relatively low, and investing becomes more attractive. The simulation in \cref{fig:invest} D also reflects this behavior. For comparison, when assuming exponential discounting (\cref{fig:invest} E and F), the value and Q function do not show preference reversal and remain constant over time. With regard to the inverse approach, \cref{fig:invest} G shows the computed gradients of the IRL objective function $F$ in \cref{eq:obj_fun_inverse} for different parameter pairs. One can observe that the gradients mostly point in the direction of the parameters used for simulation, indicated by a red cross. \Cref{fig:invest} H shows the evaluated IRL objective function $F$ for the parameter pairs, correctly displaying low values in the area close to the true parameters.

For the line problem, the computed value function is depicted in \cref{fig:line} A. Also in this task, the learned value increases over time and is high in areas close to highly rewarding states. \Cref{fig:line} B and C shows the learned Q function and policy respectively. In the beginning when the risk could be potentially high, if not close enough to the large reward, it is best to move right to collect the smaller sooner reward. With time when the risk of termination decreases, it becomes increasingly advantageous to move left to collect the larger later reward. In between, there is a time span, during which it is best to stay: The moving cost is too high for moving right and shortly turning afterward, but the risk is not low enough for being worth it to move left to the higher reward. The simulation in \cref{fig:line} D presents a sampled trajectory, showing that the subject first moves right until the maximum smaller sooner reward at $x=0.5$ is reached and stays there for some time. After the risk has decreased over time, the simulated subject finally moves right to collect the larger later reward -- her or his preferences have reversed. As for the investment problem, the value and Q function (\cref{fig:line} E and F) do not model any time-varying behavior. With regard to the IRL problem, the gradients and function values of $F$ of \cref{eq:obj_fun_inverse} in \cref{fig:line} G and H also lead closely to parameter values used for the simulations.

%% file: sections/5_conclusion.tex
\section{Conclusion}
\label{sec:conclusion}
In this work, we have proposed a method for reinforcement learning with non-exponential discount functions. The approach can be used to solve decision-making problems with an arbitrary end-time distribution and to model human discounting behavior. 
First, we have shown the conditions for which the problem is well-defined when using a hyperbolic discount function. Then, we derived a HJB-type equation providing conditions for the optimal time-dependent value function. We presented how the obtained PDE can be approximately solved using a collocation method, leading to the optimal policy. Further, we introduced an approach for the inverse problem, in which the discount function needs to be inferred given behavioral data. The application of our methods on two simulated problems led to plausible solutions, opening the way for further applications such as the use in human experiments.

\paragraph{Limitations and future work}
In our proposed methods, we assume a finite action space. While this assumption applies to many behavioral experiments, it can be a limitation for the application to classical optimal control. To extend the method to continuous control, one has to determine how the maximization problem in the HJB equation is solved. For strictly convex action costs, there have been approaches proposed that efficiently solve the maximization in the HJB equation \cite{lutter2020hjb}, under some conditions even in closed-form \cite{tassa2007least}. Further, for Lipschitz-continuous controls, it is possible to approximate the control dynamics to avoid solving the optimization problem \cite{kim2021hamilton}.
Another requirement of our method is that the model needs to be known. Model-free approaches, such as TD-error learning for general discount functions could be explored, in line with work such as \cite{alexander2010hyperbolically}.
Further, when approaching problems with large state spaces, the proposed collocation method is likely to converge slowly. For these cases, adapted collocation methods \cite{sirignano2018dgm} or advantage updating approaches \cite{baird1994reinforcement} could be considered instead. To apply our method to more advanced problems in financial engineering, such as modeling stocks with discontinuous returns \cite{merton1976option}, one has to consider general jump-diffusion processes. While an extension to these models is straightforward from a theoretical point, we here left it out to keep the focus on the handling of the discount function. Regarding our inverse reinforcement learning approach, we have assumed the states and time points of the action switches to be given. While this assumption is reasonable for many human behavior experiments, it might be interesting to learn discount functions for given discretized trajectories instead. Also, incorporating an extended timing model of human decisions \cite{k2014general} instead of fixed-variance Gaussian diffusion would be an interesting extension.

In the future, we are planning to apply our proposed methods in human experiments to get new insights to human discounting behavior.
Characterizing individual human subjects by analyzing their behavior comes with the risk to be used with negative social impact. This matter can be counteracted by collecting only anonymized data for the application of our method.

%% file: sections/6_checklist.tex
\newpage
\section*{Checklist}

\begin{enumerate}

\item For all authors...
\begin{enumerate}
  \item Do the main claims made in the abstract and introduction accurately reflect the paper's contributions and scope?
    \answerYes{All contributions are provided in \cref{sec:rl_nonexp} with applications to experiments in \cref{sec:experiments}}.
  \item Did you describe the limitations of your work?
    \answerYes{See \cref{sec:conclusion}.}
  \item Did you discuss any potential negative societal impacts of your work?
    \answerYes{See \cref{sec:conclusion}.}
  \item Have you read the ethics review guidelines and ensured that your paper conforms to them?
    \answerYes{}
\end{enumerate}

\item If you are including theoretical results...
\begin{enumerate}
  \item Did you state the full set of assumptions of all theoretical results?
    \answerYes{In the intro text in \cref{sec:rl_nonexp}.}
    \item Did you include complete proofs of all theoretical results?
    \answerYes{Complete proofs are provided in \cref{app:convergence} and \cref{app:derivation_hjb}}.
\end{enumerate}

\item If you ran experiments...
\begin{enumerate}
  \item Did you include the code, data, and instructions needed to reproduce the main experimental results (either in the supplemental material or as a URL)?
    \answerYes{See supplementary material.}
  \item Did you specify all the training details (e.g., data splits, hyperparameters, how they were chosen)?
    \answerYes{see \cref{app:hyperparams}}
    \item Did you report error bars (e.g., with respect to the random seed after running experiments multiple times)?
    \answerYes{See error bars for learned Q functions in \cref{subsec:exp_results}}
    \item Did you include the total amount of compute and the type of resources used (e.g., type of GPUs, internal cluster, or cloud provider)?
    \answerYes{See \cref{app:hyperparams}.}
\end{enumerate}

\item If you are using existing assets (e.g., code, data, models) or curating/releasing new assets...
\begin{enumerate}
  \item If your work uses existing assets, did you cite the creators?
    \answerYes{Pytorch was used and referenced.}
  \item Did you mention the license of the assets?
    \answerYes{see \cref{app:hyperparams}.}
  \item Did you include any new assets either in the supplemental material or as a URL?
    \answerYes{The code is provided in the supplementary material.}
  \item Did you discuss whether and how consent was obtained from people whose data you're using/curating?
    \answerNA{We have not used data of other people.}
  \item Did you discuss whether the data you are using/curating contains personally identifiable information or offensive content?
    \answerNA{We have not used data of other people.}
\end{enumerate}

\item If you used crowdsourcing or conducted research with human subjects...
\begin{enumerate}
  \item Did you include the full text of instructions given to participants and screenshots, if applicable?
    \answerNA{}
  \item Did you describe any potential participant risks, with links to Institutional Review Board (IRB) approvals, if applicable?
    \answerNA{}
  \item Did you include the estimated hourly wage paid to participants and the total amount spent on participant compensation?
    \answerNA{}
\end{enumerate}

\end{enumerate}

\newpage

%% file: sections/7_appendix.tex
\normalsize
\appendix
\title{Appendix\\\normalfont{Reinforcement Learning\\ with Non-Exponential Discounting}}
\author{}
\maketitlenew

\section{Convergence proof for the value function under hyperbolic discounting}
\label{app:convergence}
In the following, we assume a hyperbolic survival function as presented in \cref{eq:hyp_sf}, i.e.,
\begin{align*}
    S(t; \alpha, \beta) = \frac{1}{(\frac{t}{\beta} + 1)^{\alpha}}.
\end{align*}

\newcommand{\rsup}{r_\textrm{sup}}
\newcommand{\rinf}{r_\textrm{inf}}

\paragraph{Part I}
\textit{If the reward function $\SR(\Sx, \Su, t)$ is bounded above for all $(\Sx, \Su, t) \in \Sxset \times \Suset \times \Rplus$, and $\alpha_0 > 1$, the value function defined in equation \cref{eq:method_v_fun} is well-defined.}

We denote the supremum of the reward function $\SR(\Sx, \Su, t)$ for all $(\Sx, \Su, t) \in \Sxset \times \Suset \times \Rplus$ by $\rsup$. We find
\begin{align*}
    V^*(\Sx, t) &= \max_{\Su_{[t, \infty)}} \E \left[ \int_t^\infty \frac{S(\Sti)}{S(t)} \, \SR(\SX(\Sti), \Su(\Sti), \Sti) \, \mathrm{d}\Sti \given[\Big] \SX(t) = \Sx \right] \\
    &\leq \int_t^\infty \frac{S(\Sti)}{S(t)} \, \rsup \, \mathrm{d}\Sti \\
    &= \frac{\rsup}{S(t)} \int_t^\infty S(\Sti) \, \mathrm{d}\Sti \\
    &= \frac{\rsup}{S(t)} \int_t^\infty \frac{1}{\left(\frac{\Sti}{\beta} + 1\right)^{\alpha}}  \, \mathrm{d}\Sti \\
    &\leq \frac{\rsup}{S(t)} \int_t^\infty \frac{1}{\left(\frac{\Sti}{\beta}\right)^{\alpha}}  \, \mathrm{d}\Sti \\
    &= \frac{\beta^\alpha \, \rsup}{S(t)} \int_t^\infty \frac{1}{\Sti^{\alpha}}  \, \mathrm{d}\Sti \\
    &= \frac{\beta^\alpha \, \rsup}{S(t)} \left[ \frac{\tau^{1-\alpha}}{1-\alpha} \right]_{\tau = t}^\infty\\
    &= \frac{\beta^\alpha \, \rsup}{S(t) \, (1-\alpha)} \left[ \tau^{1-\alpha} \right]_{\tau = t}^\infty,
\end{align*}
which is finite for $\alpha > 1$.

\paragraph{Part II}

\textit{If $\SR(\Sx, \Su, t)$ is bounded below for all $(\Sx, \Su, t) \in \Sxset \times \Suset \times \Rplus$, and $\alpha_0 \leq 1$, the value function defined in equation \cref{eq:method_v_fun} is not well-defined.}

We denote the infimum of the reward function $\SR(\Sx, \Su, t)$ for all $(\Sx, \Su, t) \in \Sxset \times \Suset \times \Rplus$ by $\rinf$. We find
\begin{align*}
    V^*(\Sx, t) &= \max_{\Su_{[t, \infty)}} \E \left[ \int_t^\infty \frac{S(\Sti)}{S(t)} \, \SR(\SX(\Sti), \Su(\Sti), \Sti) \, \mathrm{d}\Sti \given[\Big] \SX(t) = \Sx \right] \\
    &\geq \int_t^\infty \frac{S(\Sti)}{S(t)} \, \rinf \, \mathrm{d}\Sti \\
    &= \frac{\rinf}{S(t)} \int_t^\infty S(\Sti) \, \mathrm{d}\Sti \\
    &= \frac{\rinf}{S(t)} \int_t^\infty \frac{1}{\left(\frac{\Sti}{\beta} + 1\right)^{\alpha}} \, \mathrm{d}\Sti \\
    &= \frac{\rinf}{S(t)} \int_t^\infty \frac{1}{\left(\frac{\Sti + \beta}{\beta}\right)^{\alpha}} \, \mathrm{d}\Sti \\
    &= \frac{\beta^\alpha \, \rinf}{S(t)} \int_t^\infty \frac{1}{( \Sti + \beta)^\alpha} \, \mathrm{d}\Sti \\
    &= \frac{\beta^\alpha \, \rinf}{S(t)} \int_{t+\beta}^\infty \frac{1}{\Sti^\alpha} \, \mathrm{d}\Sti \\
    &= \frac{\beta^\alpha \, \rinf}{S(t)} \left[ \frac{1}{\Sti^\alpha} \right]_{\Sti=t+\beta}^\infty \\
    &= \frac{\beta^\alpha \, \rinf}{S(t)} \left[ \frac{\tau^{1-\alpha}}{1-\alpha} \right]_{\tau = t+\beta}^\infty\\
    &= \frac{\beta^\alpha \, \rinf}{S(t) \, (1-\alpha)} \left[ \tau^{1-\alpha} \right]_{\tau = t+\beta}^\infty,
\end{align*}
in which the integral diverges for $\alpha \leq 1$.

\section{Full derivation of the HJB equation}
\label{app:derivation_hjb}

In this section, we provide a full derivation for the HJB equation introduced in \cref{sec:hjb_general}. 
We start with the value function defined in \cref{eq:method_v_fun}, i.e.,
\begin{align*}
    V^*(\Sx, t) = \max_{\Su_{[t, \infty)}} \E \left[ \int_t^\infty \frac{S(\Sti)}{S(t)} \SR(\SX(\Sti), \Su(\Sti), \Sti) \, \mathrm{d}\Sti \given[\Big] \SX(t) = \Sx \right].
\end{align*}
First, we split the integral into two terms and obtain
\begin{align*}
    V^*(\Sx, t) = \max_{\Su_{[t, t+ \Delta t]}} \E &\left[ \int_t^{t+\Delta t} \frac{S(\Sti)}{S(t)} \SR(\SX(\Sti), \Su(\Sti), \Sti) \, \td\Sti \right.\\
    &+\left. \int_{t+\Delta t}^{\infty} \frac{S(\Sti)}{S(t)} \SR(\SX(\Sti), \Su(\Sti), \Sti) \, \td\Sti \given[\Big] \SX(t) = \Sx \right].
\end{align*}
By identifying the second term as the value function of state $x(t+\Sdt)$ at time $t+\Sdt$, we obtain the recursive formulation
\begin{align*}
    V^*(\Sx, t) 
    = \max_{\Su_{[t, t+ \Delta t]}} \E &\left[ \int_t^{t+\Delta t} \frac{S(\Sti)}{S(t)} \SR(\SX(\Sti), \Su(\Sti), \Sti) \, \td\Sti \right.\\
    &+\left. \frac{S(t + \Delta t)}{S(t)} V^*(\SX(t+ \Delta t), t + \Delta t) \given[\Big] \SX(t) = \Sx \right].
\end{align*}
Consider a small $\Delta t$, then the first term evaluates to
\begin{align*}
    \int_t^{t+\Delta t} \frac{S(\Sti)}{S(t)} \SR(\SX(\Sti), \Su(\Sti), \Sti) \, \mathrm{d}\Sti = \SR(\SX(t), \Su(t), t) \cdot \Delta t + \littleO(\Sdt).
\end{align*}
For the second term, we apply a Taylor expansion and get
\begin{align*}
    V^*(\SX(t+ \Sdt)&, t + \Sdt) 
    = V^*(\SX(t), t) + \int_t^{t+\Sdt} \frac{\mathrm{d}}{\mathrm{d} \Sti} V^*(\SX(\Sti), \Sti) \, \mathrm{d}\Sti + \littleO(\Sdt) \\
    &= V^*(\SX(t), t) + \int_t^{t+\Sdt} \Vx^*(\SX(\Sti), \Sti) \, \mathrm{d}\SX(\Sti) + \int_t^{t+\Sdt} \Vt^*(\SX(\Sti), \Sti) \, \mathrm{d}\Sti + \littleO(\Sdt).
\end{align*}
Here, the second term can be evaluated using It\^{o}'s formula as
\begin{align*}
    \int_t^{t+\Sdt} \Vx^*(\SX(\Sti), \Sti) \, &\mathrm{d}\SX(\Sti) = \int_t^{t+\Sdt} \Vx^*(\SX(\Sti), \Sti) \, f(\SX(\Sti), \Su(\Sti), \Sti)\, \td\Sti\\
    &+ \int_t^{t+\Sdt} \frac{1}{2} \tr{\left\{\Vxx^*(\SX(\Sti), \Sti) \, \SG(\SX(\Sti), \Su(\Sti), \Sti) \, \SG(\SX(\Sti), \Su(\Sti), \Sti)^T \right\}} \, \td\Sti \\
    &+ \int_t^{t+\Sdt} \Vx^*(\SX(\Sti), \Sti) \, \SG(\SX(\Sti), \Su(\Sti), \Sti) \, \td\SW(\Sti) + \littleO(\Sdt).
\end{align*}
Plugging in these terms into the equation above and dividing both sides by $\Sdt$ yields
\begin{align*}
    \frac{1-\frac{S(t+\Sdt)}{S(t)}}{\Sdt} &V^*(\SX(t), t) = \max_{\Su_{[t, t+ \Delta t]}} \E \left[ \frac{1}{\Sdt} \int_t^{t+\Delta t} \frac{S(\Sti)}{S(t)} \SR(\SX(\Sti), \Su(\Sti), \Sti) \, \mathrm{d}\Sti \right.\\
    & + \frac{1}{\Sdt} \int_t^{t+\Sdt} \Vx^*(\SX(\Sti), \Sti) \, f(\SX(\Sti), \Su(\Sti), \Sti)\, \td\Sti + \int_t^{t+\Sdt} \Vt^*(\SX(\Sti), \Sti) \, \td\Sti \\
    & + \frac{1}{\Sdt} \int_t^{t+\Sdt} \frac{1}{2} \tr{\left\{\Vxx^*(\SX(\Sti), \Sti) \, \SG(\SX(\Sti), \Su(\Sti), \Sti) \, \SG(\SX(\Sti), \Su(\Sti), \Sti)^T \right\}} \, \td\Sti\\
    & + \frac{1}{\Sdt} \left. \int_t^{t+\Sdt} \Vx^*(\SX(\Sti), \Sti) \, \SG(\SX(\Sti), \Su(\Sti), \Sti) \, \td\SW(\Sti)
     + \frac{\littleO(\Sdt)}{\Sdt} \given[\Big] \SX(t) = \Sx \right].  
\end{align*}
The factor on the l.h.s. in the limit $\Sdt \to 0$ can be recognized to be the hazard rate (cf.\ \cref{eq:surv_analysis}),
\begin{align*}
    \lim_{\Sdt \to 0} \frac{1-\frac{S(t+\Sdt)}{S(t)}}{\Sdt} = \lim_{\Sdt \to 0} \frac{1}{\Sdt} \frac{S(t) - S(t+ \Sdt)}{\Sdt} = \alpha(t).
\end{align*}
Taking the limit $\Sdt \to 0$ on both sides and calculating the expectation w.r.t.\ $\SW(t)$,
we obtain the HJB equation
\begin{align*}
    \alpha(t) V^*(\Sx, t) = \max_{\Su} \  &\left[\SR(\Sx, \Su, t)+ \Vt^*(\Sx, t) + \Vx^*(\Sx, t) \, f(\Sx, \Su, t) \right. \\
    & \quad + \frac{1}{2} \left.\tr{\left\{\Vxx^*(\Sx, t) \, \SG(\Sx, \Su, t) \, \SG(\Sx, \Su, t)^T \right\}} \right].
\end{align*}

\section{Bellman equation for discrete time}
We consider the discrete-time setting, in which the objective is given as
\label{app:bellman_discrete_time}
\begin{align*}
    J\left(\Su_0, \Su_1, \dots\right) = \E \left[ \displaystyle\sum_{\tau=0}^\infty S(\Sti) \, \SR(\SX_\Sti, \Su_\Sti, \Sti) \right].
\end{align*}
As in the continuous-time case, we can define the value function as
\begin{align*}
    V(\Sx, t) &= \max_{\Su_t, \Su_{t+1}, \dots} \E \left[ \displaystyle\sum_{\tau=t}^\infty \frac{S(\Sti)}{S(t)} \, \SR(\SX_\Sti, \Su_\Sti, \Sti) \given[\Big] \SX_t = \Sx \right].
\end{align*}
By identifying the recursive definition of the value function and evaluating terms, we obtain the Bellman equation
\begin{align*}
    V(\Sx, t) &= \max_{\Su} \left\{ R(\Sx, \Su, t) + \lambda(t) \E \left[ V(\SX_{t+1}, t+1) \given \SX_{t} = \Sx \right] \right\},
\end{align*}
with $\lambda(t) = S(t+1)/S(t)$ being the hazard probability at time t.

\section{Value function approximation and collocation method}
\label{app:app_collocation}
In the collocation method in \cref{alg:collocation}, we need to sample random states $\hat{\Sx}_i$ and time points $\hat{t}_i$ for minimizing $\sum_i E(\V^\psi, \hat{\Sx}_i, \hat{t}_i) ^ 2$. If we assume a bounded state space $\Sxset \in \R$, we can sample $\hat{\Sx}_i$ uniformly from this space. The time points $\hat{t}_i \in \Rplus$ can be sampled from an exponential distribution. To do so, we first draw $\hat{y}_i \sim \UniformDis(0, 1)$ and compute $\hat{t}_i = - \log(1-\hat{y}_i) / \lambda$. To feed a normalized value of time into the network, we use $\hat{y}_i$ instead of $\hat{t}_i$ as input to the network. We denote the value function network depending on $y$ by $\tilde{V}(\Sx, y)$. Given a specific time value $t$, we can compute its representation via $y(t) = 1 - \exp(-\lambda t)$.

When computing the partial derivative $\Vt$, we have to take this reparametrization into account. By the chain rule, we find
\begin{align*}
    \Vt (\Sx, t) = \tilde{V}_y (\Sx, t) \, y_t (t),
\end{align*}
for which we have with the chosen parametrization
\begin{align*}
    y_t (t) = \lambda \exp(-\lambda t) .
\end{align*}

In general, there are multiple solutions to the HJB equation and the encountered solution depends on the initialization of the function approximator \cite{tassa2007least, lutter2020hjb}. In other work, this problem has been dealt with by omitting stochastic terms in the first episodes of training or annealing the discount factor \cite{tassa2007least, lutter2020hjb, alt2020pomdps}. We adopt the second approach and move from short to far-sighted discounting to converge to the desired solution. For hyperbolic discounting, we initially add an offset to $\alpha_0$, leading to a high expected hazard rate. Over time, we decrease the offset to converge to the desired solution.

\section{Experiments}
\label{app:experiments}

\subsection*{Investment problem}
\begin{itemize}
\item
State space $\Sxset = [0, 1] \times [0, 1]$, modeling account balance and interest rate, i.e., $\Sx = [x_b, x_i]$
\item
Action space $\Suset = \{\textit{spend}, \textit{invest}\}$
\item
Dynamics model
\begin{align*}
    f(\Sx, \Su) &{}= 
    \begin{cases}
    [0, 0]^T   & \quad \text{if } \Su \text{ is \textit{spend}}\\
    [0.1, 0]^T & \quad \text{if } \Su \text{ is \textit{invest}}
    \end{cases}\\
    G(\Sx, \Su) &{}=  
    \begin{pmatrix}
    0 & 0\\
    0 & 0.01
    \end{pmatrix}
\end{align*}
\item
Reward function
\begin{align*}
    \SR(\Sx, \Su) &= \SR^\Sx(\Sx) + \SR^\Su(\Su)\\
    \SR^\Sx([x_b, x_i]) &= x_b \cdot x_i\\
    \SR^\Su(\Su) &= 
    \begin{cases}
        0.1 & \quad \text{if } \Su \text{ is \textit{spend}}\\
        0   & \quad \text{if } \Su \text{ is \textit{invest}}
    \end{cases}
\end{align*}
\item
Initial belief of hazard rate 
$\alpha_0 = 3, \beta_0 = 1$ (visualized in \cref{fig:curves})
\end{itemize}

\subsection*{Line problem}
\begin{figure}[t!]
    \centering
    \includegraphics[width=\textwidth]{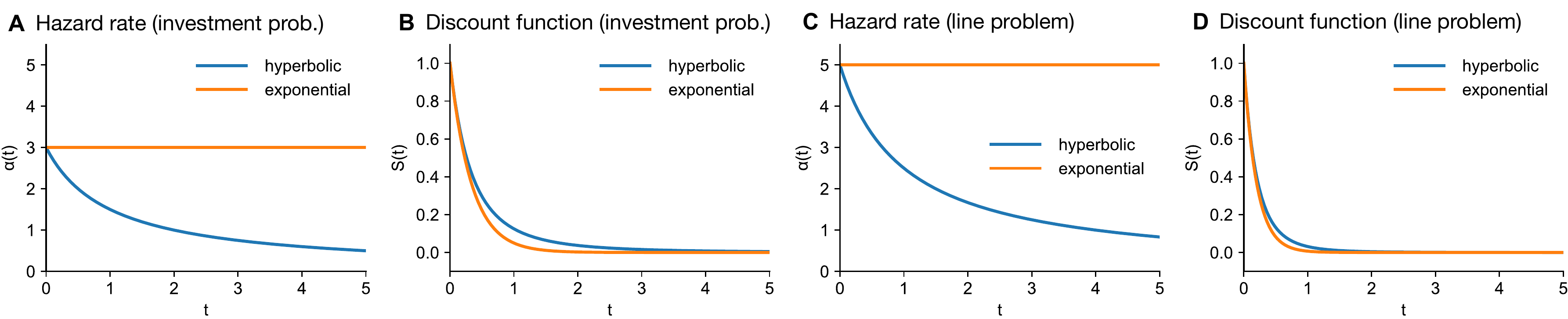}
    \caption{\textbf{Hazard rates and Discounting functions.} \textbf{A} Expected hazard rate for the investment problem. For hyperbolic discounting, the expected risk of termination is decreasing over time, while for exponential discounting, the hazard rate is constant. \textbf{B} Expected discount function for the investment problem in comparison to an exponential discount function. \textbf{C} Expected hazard rate for the line problem for hyperbolic discounting in comparison with the constant hazard rate when applying exponential discounting \textbf{D} Expected discount function for the line problem in comparison to an exponential discount function.}
    \label{fig:curves}
\end{figure}

\begin{itemize}
\item
State space $\Sxset = [-1, 1]$
\item
Action space $\Suset = \{\textit{left}, \textit{stay}, \textit{right}\}$
\item Dynamics model
\begin{align*}
    f(\Sx, \Su) &{}= 
    \begin{cases}
    -1  & \quad \text{if } \Su \text{ is \textit{left}}\\
    0   & \quad \text{if } \Su \text{ is \textit{stay}}\\
    1   & \quad \text{if } \Su \text{ is \textit{right}}
    \end{cases}\\
    G(\Sx, \Su) &{}=
    \begin{cases}
        0.05 & \quad \text{if } \Su \in \{\textit{left}, \textit{right}\}\\
        0   & \quad \text{if } \Su \text{ is \textit{stay}}
    \end{cases}
\end{align*}

\item
Reward function

\begin{minipage}{\linewidth} 
\begin{wrapfigure}{r}{0.3\textwidth} 
\vspace{10pt}
\includegraphics[width=0.3\textwidth]{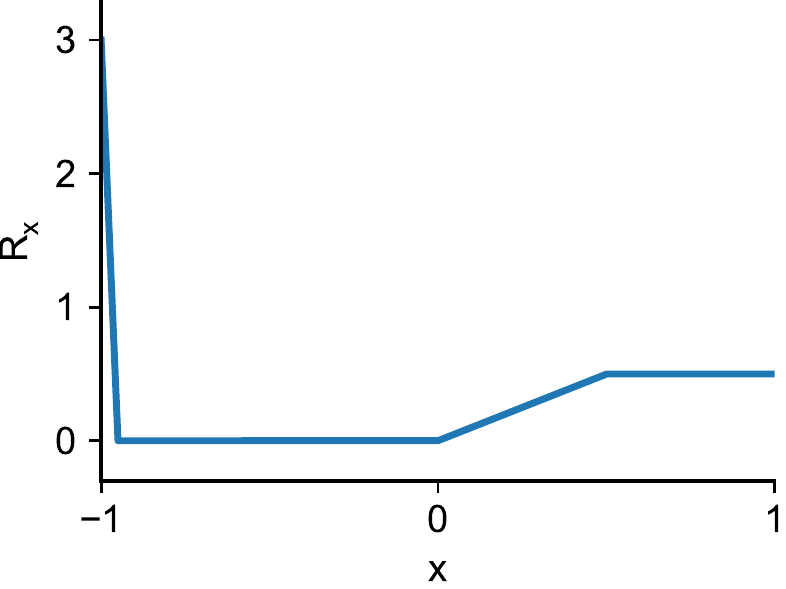} 
\end{wrapfigure} 
\begin{align*}
    \SR(\Sx, \Su) &= \SR^\Sx(\Sx) + \SR^\Su(\Su)\\
    \SR^\Sx(\Sx) &=
    \begin{cases}
        0.5     & \quad \text{if } \Sx \geq 0.5\\
        \Sx     & \quad \text{if } 0 \leq \Sx < 0.5\\
        0       & \quad \text{if } -0.95 \leq \Sx < 0 \\
        -60 \Sx -57 & \quad \text{if } \Sx < 0.95 \\
    \end{cases}\\
    \SR^\Su(\Su) &=
    \begin{cases}
        0.1 & \quad \text{if } \Su \in \{\textit{left}, \textit{right}\}\\
        0   & \quad \text{if } \Su \text{ is \textit{stay}}
    \end{cases}
\end{align*}
\end{minipage}
\item
Initial belief of hazard rate
$\alpha_0 = 5, \beta_0 = 1$ (visualized in \cref{fig:curves})
\end{itemize}

\section{Hyperparameters, implementation, and computing resources}
\label{app:hyperparams}

Throughout the experiments, we have used the following hyperparameters:
\begin{itemize}
\item
The neural networks are parametrized as
\begin{verbatim}
    layers = (nn.Linear(input_dim, layer_size),
              nn.Sigmoid(),
              nn.Linear(layer_size, layer_size),
              nn.Sigmoid(),
              nn.Linear(layer_size, output_dim))
    model = nn.Sequential(*layers)
\end{verbatim}
\item For the neural network representing $\V$, we used
\begin{verbatim}
    input_dim = x_dim + 1
    output_dim = 1
\end{verbatim}
\item For the neural network representing $\V_\theta$, we used
\begin{verbatim}
    input_dim = x_dim + 1
    output_dim = theta_dim
\end{verbatim}
\item We set $\lambda = 0.2$.
\item For the collocation method, we used 10.000 samples in each iteration and 125.000 episodes for the investment problem and 100.000 episodes for the line problem. The initial offset of $\alpha_0$ was set to 50 and linearly decreased to zero over 50.000 episodes.
\item We used Adam optimizer with learning rate 0.003.
\item For the runs with exponential discounting, the mean of the initial belief over the hazard rate was taken for $\lambda$, i.e., 3 for the investment problem and 5 for the line problem.
\end{itemize}

More information about Implementation and computing resources:
\begin{itemize}
    \item
    Methods were implemented in Python using the PyTorch framework \cite{paszke2019pytorch}, which has been published under a BSD license.
    \item
    Resources used:
    Intel\textsuperscript{\textregistered} Xeon\textsuperscript{\textregistered} Platinum 9242 Processor, using 8 cores per run.
    \item Network training took \textasciitilde50 min.\ for the investment problem and \textasciitilde30 min.\ for the line problem.
\end{itemize}

\section{Derivation of the hyperbolic discount function as uncertainty over the constant hazard rate}
\label{app:hyperbolic_unc_hazard}
We assume $P(T > t \given \lambda) = \exp(-\lambda t)$ and a belief $\lambda \sim \GamDis(\lambda; \alpha, \beta)$. For the expected survival function, we calculate
\begin{align*}
    S(t) &= \int_{\lambda} \exp{(-\lambda t)} p\left(\lambda\right) \, \mathrm{d}\lambda\\
    &= \int_{\lambda} \exp{(-\lambda t)} \frac{\beta^\alpha \lambda^{\alpha -1} \exp(-\beta \lambda)}{\Gamma(\alpha)} \, \mathrm{d}\lambda\\
    &= \int_{\lambda} \frac{\beta^\alpha \lambda^{\alpha-1} \exp (-(\beta + t)\lambda) }{\Gamma(\alpha)} \, \mathrm{d}\lambda\\
    &= \frac{\beta^\alpha}{(\beta + t)^\alpha} \int_{\lambda} \mathrm{Gamma}(\lambda; \alpha, \beta + t) \, \mathrm{d}\lambda\\
    &= \frac{1}{\left(\frac{t}{\beta} + 1\right)^\alpha}.
\end{align*}

\section{Interpretation of the discount factor as transition to terminal state}
A Markov decision process (MDP) with discounting can be converted to an MDP without discounting by adding an additional terminal state $\Upsilon$ \cite{bertsekas2019reinforcement}. From each state with a certain probability $\gamma$, one transitions to the terminal state, and the remaining transition probabilities are renormalized. At the terminal state there is no possibility to transition to any other state and a reward of zero is given. In continuous time, the same formalization can be applied, but we consider a rate instead at which one transitions to the terminal state. Further, we assume in the following that the rate depends on time and denote it by $\lambda(t)$. The probability to be in the terminal states at time $\Upsilon$ is given by the cumulative distribution function (CDF),
\begin{align*}
    P(\SX(t) = \Upsilon) &= P(T < t).
\end{align*}
The probability of not having terminated yet is given by the complementary cumulative distribution function (CCDF),
\begin{align*}
    P(\SX(t) \neq \Upsilon) &= 1 - P(\SX(t) = \Upsilon)\\
    &= P(T \geq t)\\
    &= S(t),
\end{align*}
which is equal to the discount function.

For a constant termination rate, one obtains the CDF and CCDF of the exponential distribution, respectively:
\begin{align*}
    P(\SX(t) = \Upsilon) &= \lambda \int_0^t \exp(-\lambda \Sti) \, \td \Sti\\
    &= 1 - \exp(-\lambda t)\\\\
    P(\SX(t) \neq \Upsilon) &= 1 - P(\SX(t) = \Upsilon)\\
    &= \exp(-\lambda t)
\end{align*}

%% file: main.bbl
\begin{thebibliography}{75}
\providecommand{\natexlab}[1]{#1}
\providecommand{\url}[1]{\texttt{#1}}
\expandafter\ifx\csname urlstyle\endcsname\relax
  \providecommand{\doi}[1]{doi: #1}\else
  \providecommand{\doi}{doi: \begingroup \urlstyle{rm}\Url}\fi

\bibitem[O'Donoghue and Rabin(2000)]{o2000economics}
Ted O'Donoghue and Matthew Rabin.
\newblock The economics of immediate gratification.
\newblock \emph{{Journal of Behavioral Decision Making}}, 13\penalty0
  (2):\penalty0 233--250, 2000.

\bibitem[Frederick et~al.(2002)Frederick, Loewenstein, and
  O'donoghue]{frederick2002time}
Shane Frederick, George Loewenstein, and Ted O'donoghue.
\newblock Time discounting and time preference: A critical review.
\newblock \emph{{Journal of Economic Literature}}, 40\penalty0 (2):\penalty0
  351--401, 2002.

\bibitem[McKerchar et~al.(2009)McKerchar, Green, Myerson, Pickford, Hill, and
  Stout]{mckerchar2009comparison}
Todd~L McKerchar, Leonard Green, Joel Myerson, T~Stephen Pickford, Jade~C Hill,
  and Steven~C Stout.
\newblock A comparison of four models of delay discounting in humans.
\newblock \emph{{Behavioural Processes}}, 81\penalty0 (2):\penalty0 256--259,
  2009.

\bibitem[Strotz(1955)]{strotz1955myopia}
Robert~Henry Strotz.
\newblock Myopia and inconsistency in dynamic utility maximization.
\newblock \emph{{The Review of Economic Studies}}, 23\penalty0 (3):\penalty0
  165--180, 1955.

\bibitem[Thaler(1981)]{thaler1981some}
Richard Thaler.
\newblock Some empirical evidence on dynamic inconsistency.
\newblock \emph{{Economics Letters}}, 8\penalty0 (3):\penalty0 201--207, 1981.

\bibitem[Mazur(1987)]{mazur1987adjusting}
James~E Mazur.
\newblock An adjusting procedure for studying delayed reinforcement.
\newblock \emph{{Quantitative Analyses of Behavior}}, 5:\penalty0 55--73, 1987.

\bibitem[Fisher(1930)]{fisher1930theory}
Irving Fisher.
\newblock \emph{Theory of interest: as determined by impatience to spend income
  and opportunity to invest it}.
\newblock Augustusm Kelly Publishers, 1930.

\bibitem[Dasgupta and Maskin(2005)]{dasgupta2005uncertainty}
Partha Dasgupta and Eric Maskin.
\newblock Uncertainty and hyperbolic discounting.
\newblock \emph{{American Economic Review}}, 95\penalty0 (4):\penalty0
  1290--1299, 2005.

\bibitem[Takahashi(2005)]{takahashi2005loss}
Taiki Takahashi.
\newblock Loss of self-control in intertemporal choice may be attributable to
  logarithmic time-perception.
\newblock \emph{{Medical Hypotheses}}, 65\penalty0 (4):\penalty0 691--693,
  2005.

\bibitem[Ray and Bossaerts(2011)]{ray2011positive}
Debajyoti Ray and Peter Bossaerts.
\newblock Positive temporal dependence of the biological clock implies
  hyperbolic discounting.
\newblock \emph{{Frontiers in Neuroscience}}, 5:\penalty0 2, 2011.

\bibitem[Sozou(1998)]{sozou1998hyperbolic}
Peter~D Sozou.
\newblock On hyperbolic discounting and uncertain hazard rates.
\newblock \emph{{Proceedings of the Royal Society of London. Series B:
  Biological Sciences}}, 265\penalty0 (1409):\penalty0 2015--2020, 1998.

\bibitem[Sutton and Barto(2018)]{sutton2018reinforcement}
Richard~S Sutton and Andrew~G Barto.
\newblock \emph{Reinforcement learning: An introduction}.
\newblock {MIT Press}, 2018.

\bibitem[Bertsekas(2019)]{bertsekas2019reinforcement}
Dimitri Bertsekas.
\newblock \emph{Reinforcement learning and optimal control}.
\newblock Athena Scientific, 2019.

\bibitem[Doya(2000)]{doya2000reinforcement}
Kenji Doya.
\newblock Reinforcement learning in continuous time and space.
\newblock \emph{{Neural Computation}}, 12\penalty0 (1):\penalty0 219--245,
  2000.

\bibitem[Puterman(1990)]{puterman1990markov}
Martin~L Puterman.
\newblock Markov decision processes.
\newblock \emph{Handbooks in operations research and management science},
  2:\penalty0 331--434, 1990.

\bibitem[Alexander and Brown(2010)]{alexander2010hyperbolically}
William~H Alexander and Joshua~W Brown.
\newblock Hyperbolically discounted temporal difference learning.
\newblock \emph{{Neural Computation}}, 22\penalty0 (6):\penalty0 1511--1527,
  2010.

\bibitem[Fedus et~al.(2019)Fedus, Gelada, Bengio, Bellemare, and
  Larochelle]{fedus2019hyperbolic}
William Fedus, Carles Gelada, Yoshua Bengio, Marc~G Bellemare, and Hugo
  Larochelle.
\newblock Hyperbolic discounting and learning over multiple horizons.
\newblock \emph{arXiv preprint arXiv:1902.06865}, 2019.

\bibitem[Stratonovich(1968)]{stratonovich1968conditional}
Ruslan~L Stratonovich.
\newblock Conditional {M}arkov processes and their application to the theory of
  optimal control.
\newblock 1968.

\bibitem[Kushner and Dupuis(2001)]{kushner2001numerical}
Harold~J Kushner and Paul~G Dupuis.
\newblock \emph{Numerical methods for stochastic control problems in continuous
  time}, volume~24.
\newblock Springer Science \& Business Media, 2001.

\bibitem[Fleming and Soner(2006)]{fleming2006controlled}
Wendell~H Fleming and Halil~Mete Soner.
\newblock \emph{Controlled {M}arkov processes and viscosity solutions},
  volume~25.
\newblock Springer Science \& Business Media, 2006.

\bibitem[Pontryagin(1987)]{pontryagin1987mathematical}
Lev~Semenovich Pontryagin.
\newblock \emph{Mathematical theory of optimal processes}.
\newblock {CRC Press}, 1987.

\bibitem[Vamvoudakis and Lewis(2010)]{vamvoudakis2010online}
Kyriakos~G Vamvoudakis and Frank~L Lewis.
\newblock Online actor–critic algorithm to solve the continuous-time infinite
  horizon optimal control problem.
\newblock \emph{Automatica}, 46\penalty0 (5):\penalty0 878--888, 2010.

\bibitem[Baird(1994)]{baird1994reinforcement}
Leemon~C Baird.
\newblock Reinforcement learning in continuous time: Advantage updating.
\newblock In \emph{IEEE International Conference on Neural Networks}, volume~4,
  pages 2448--2453. IEEE, 1994.

\bibitem[Jacobson(1968)]{jacobson1968new}
David~H Jacobson.
\newblock New second-order and first-order algorithms for determining optimal
  control: A differential dynamic programming approach.
\newblock \emph{{Journal of Optimization Theory and Applications}}, 2\penalty0
  (6):\penalty0 411--440, 1968.

\bibitem[Tassa et~al.(2007)Tassa, Erez, and Smart]{tassa2007receding}
Yuval Tassa, Tom Erez, and William Smart.
\newblock Receding horizon differential dynamic programming.
\newblock In \emph{{Advances in Neural Information Processing Systems}},
  volume~20. Curran Associates, Inc., 2007.

\bibitem[Kappen(2005)]{kappen2005path}
Hilbert~J Kappen.
\newblock Path integrals and symmetry breaking for optimal control theory.
\newblock \emph{{Journal of Statistical Mechanics: Theory and Experiment}},
  2005\penalty0 (11):\penalty0 P11011, 2005.

\bibitem[Theodorou et~al.(2010)Theodorou, Buchli, and
  Schaal]{theodorou2010generalized}
Evangelos Theodorou, Jonas Buchli, and Stefan Schaal.
\newblock A generalized path integral control approach to reinforcement
  learning.
\newblock \emph{Journal of Machine Learning Research}, 11:\penalty0 3137--3181,
  2010.

\bibitem[Simpkins and Todorov(2009)]{simpkins2009practical}
Alex Simpkins and Emanuel Todorov.
\newblock Practical numerical methods for stochastic optimal control of
  biological systems in continuous time and space.
\newblock In \emph{{IEEE Symposium on Adaptive Dynamic Programming and
  Reinforcement Learning}}, pages 212--218. IEEE, 2009.

\bibitem[Tassa and Erez(2007)]{tassa2007least}
Yuval Tassa and Tom Erez.
\newblock Least squares solutions of the {HJB} equation with neural network
  value-function approximators.
\newblock \emph{{IEEE Transactions on Neural Networks}}, 18\penalty0
  (4):\penalty0 1031--1041, 2007.

\bibitem[Sirignano and Spiliopoulos(2018)]{sirignano2018dgm}
Justin Sirignano and Konstantinos Spiliopoulos.
\newblock {DGM}: {A} deep learning algorithm for solving partial differential
  equations.
\newblock \emph{{Journal of Computational Physics}}, 375:\penalty0 1339--1364,
  2018.

\bibitem[Lutter et~al.(2020)Lutter, Belousov, Listmann, Clever, and
  Peters]{lutter2020hjb}
Michael Lutter, Boris Belousov, Kim Listmann, Debora Clever, and Jan Peters.
\newblock {HJB} optimal feedback control with deep differential value functions
  and action constraints.
\newblock In \emph{{Conference on Robot Learning}}, pages 640--650. PMLR, 2020.

\bibitem[Han et~al.(2018)Han, Jentzen, and Weinan]{han2018solving}
Jiequn Han, Arnulf Jentzen, and E~Weinan.
\newblock Solving high-dimensional partial differential equations using deep
  learning.
\newblock volume 115, pages 8505--8510. National Academy of Sciences, 2018.

\bibitem[Alt et~al.(2020)Alt, Schultheis, and Koeppl]{alt2020pomdps}
Bastian Alt, Matthias Schultheis, and Heinz Koeppl.
\newblock {POMDPs} in continuous time and discrete spaces.
\newblock In \emph{{Advances in Neural Information Processing Systems}},
  volume~33, pages 13151--13162. Curran Associates, Inc., 2020.

\bibitem[Thaler and Shefrin(1981)]{thaler1981economic}
Richard~H Thaler and Hersh~M Shefrin.
\newblock An economic theory of self-control.
\newblock \emph{Journal of Political Economy}, 89\penalty0 (2):\penalty0
  392--406, 1981.

\bibitem[Ainslie and Herrnstein(1981)]{ainslie1981preference}
George Ainslie and Richard~J Herrnstein.
\newblock Preference reversal and delayed reinforcement.
\newblock \emph{{Animal Learning \& Behavior}}, 9\penalty0 (4):\penalty0
  476--482, 1981.

\bibitem[Green et~al.(1994)Green, Fristoe, and Myerson]{green1994temporal}
Leonard Green, Nathanael Fristoe, and Joel Myerson.
\newblock Temporal discounting and preference reversals in choice between
  delayed outcomes.
\newblock \emph{{Psychonomic Bulletin \& Review}}, 1\penalty0 (3):\penalty0
  383--389, 1994.

\bibitem[Andersen et~al.(2014)Andersen, Harrison, Lau, and
  Rutstr{\"o}m]{andersen2014discounting}
Steffen Andersen, Glenn~W Harrison, Morten~I Lau, and E~Elisabet Rutstr{\"o}m.
\newblock Discounting behavior: A reconsideration.
\newblock \emph{{European Economic Review}}, 71:\penalty0 15--33, 2014.

\bibitem[Richards et~al.(1997)Richards, Mitchell, De~Wit, and
  Seiden]{richards1997determination}
Jerry~B Richards, Suzanne~H Mitchell, Harriet De~Wit, and Lewis~S Seiden.
\newblock Determination of discount functions in rats with an adjusting-amount
  procedure.
\newblock \emph{Journal of the Experimental Analysis of Behavior}, 67\penalty0
  (3):\penalty0 353--366, 1997.

\bibitem[Coller and Williams(1999)]{coller1999eliciting}
Maribeth Coller and Melonie~B Williams.
\newblock Eliciting individual discount rates.
\newblock \emph{Experimental Economics}, 2\penalty0 (2):\penalty0 107--127,
  1999.

\bibitem[Andersen et~al.(2008)Andersen, Harrison, Lau, and
  Rutstr{\"o}m]{andersen2008eliciting}
Steffen Andersen, Glenn~W Harrison, Morten~I Lau, and E~Elisabet Rutstr{\"o}m.
\newblock Eliciting risk and time preferences.
\newblock \emph{Econometrica}, 76\penalty0 (3):\penalty0 583--618, 2008.

\bibitem[Noor(2009)]{noor2009hyperbolic}
Jawwad Noor.
\newblock Hyperbolic discounting and the standard model: eliciting discount
  functions.
\newblock \emph{Journal of Economic Theory}, 144\penalty0 (5):\penalty0
  2077--2083, 2009.

\bibitem[Cavagnaro et~al.(2016)Cavagnaro, Aranovich, McClure, Pitt, and
  Myung]{cavagnaro2016functional}
Daniel~R Cavagnaro, Gabriel~J Aranovich, Samuel~M McClure, Mark~A Pitt, and
  Jay~I Myung.
\newblock On the functional form of temporal discounting: an optimized adaptive
  test.
\newblock \emph{{Journal of Risk and Uncertainty}}, 52\penalty0 (3):\penalty0
  233--254, 2016.

\bibitem[Chang et~al.(2019)Chang, Kim, Zhang, Pitt, and
  Myung]{chang2019modeling}
Jorge Chang, Jiseob Kim, Byoung-Tak Zhang, Mark~A Pitt, and Jay~I Myung.
\newblock Modeling delay discounting using gaussian process with active
  learning.
\newblock In \emph{{CogSci}}, pages 1479--1485, 2019.

\bibitem[Schweighofer et~al.(2006)Schweighofer, Shishida, Han, Okamoto, Tanaka,
  Yamawaki, and Doya]{schweighofer2006humans}
Nicolas Schweighofer, Kazuhiro Shishida, Cheol~E Han, Yasumasa Okamoto, Saori~C
  Tanaka, Shigeto Yamawaki, and Kenji Doya.
\newblock Humans can adopt optimal discounting strategy under real-time
  constraints.
\newblock \emph{PLoS Computational Biology}, 2\penalty0 (11):\penalty0 e152,
  2006.

\bibitem[Seinstra et~al.(2018)Seinstra, Sellitto, and
  Kalenscher]{seinstra2018rate}
Maayke~Suzanne Seinstra, Manuela Sellitto, and Tobias Kalenscher.
\newblock Rate maximization and hyperbolic discounting in human experiential
  intertemporal decision making.
\newblock \emph{Behavioral Ecology}, 29\penalty0 (1):\penalty0 193--203, 2018.

\bibitem[Howard(1963)]{howard1963semi}
Ronald~A Howard.
\newblock Semi-{M}arkovian decision processes.
\newblock \emph{Bulletin of the International Statistical Institute},
  40\penalty0 (2):\penalty0 625--652, 1963.

\bibitem[Korolyuk et~al.(1975)Korolyuk, Brodi, and Turbin]{korolyuk1975semi}
V~S Korolyuk, S~M Brodi, and A~F Turbin.
\newblock Semi-{M}arkov processes and their applications.
\newblock \emph{{Journal of Soviet Mathematics}}, 4\penalty0 (3):\penalty0
  244--280, 1975.

\bibitem[Ross(1970)]{ross1970average}
Sheldon~M Ross.
\newblock Average cost semi-{M}arkov decision processes.
\newblock \emph{{Journal of Applied Probability}}, 7\penalty0 (3):\penalty0
  649--656, 1970.

\bibitem[Bradtke and Duff(1994)]{bradtke1994reinforcement}
Steven Bradtke and Michael Duff.
\newblock Reinforcement learning methods for continuous-time {M}arkov decision
  problems.
\newblock In \emph{{Advances in Neural Information Processing Systems}},
  volume~7. MIT Press, 1994.

\bibitem[Ja{\'s}kiewicz and Nowak(2021)]{jaskiewicz2021markov}
Anna Ja{\'s}kiewicz and Andrzej~S Nowak.
\newblock Markov decision processes with quasi-hyperbolic discounting.
\newblock \emph{{Finance and Stochastics}}, 25\penalty0 (2):\penalty0 189--229,
  2021.

\bibitem[Bjork and Murgoci(2010)]{bjork2010general}
Tomas Bjork and Agatha Murgoci.
\newblock A general theory of {M}arkovian time inconsistent stochastic control
  problems.
\newblock \emph{{SSRN}}, 2010.

\bibitem[Bj{\"o}rk and Murgoci(2014)]{bjork2014theory}
Tomas Bj{\"o}rk and Agatha Murgoci.
\newblock A theory of {M}arkovian time-inconsistent stochastic control in
  discrete time.
\newblock \emph{{Finance and Stochastics}}, 18\penalty0 (3):\penalty0 545--592,
  2014.

\bibitem[Grenadier and Wang(2007)]{grenadier2007investment}
Steven~R Grenadier and Neng Wang.
\newblock Investment under uncertainty and time-inconsistent preferences.
\newblock \emph{Journal of Financial Economics}, 84\penalty0 (1):\penalty0
  2--39, 2007.

\bibitem[Zou et~al.(2014)Zou, Chen, and Wedge]{zou2014finite}
Ziran Zou, Shou Chen, and Lei Wedge.
\newblock Finite horizon consumption and portfolio decisions with stochastic
  hyperbolic discounting.
\newblock \emph{Journal of Mathematical Economics}, 52:\penalty0 70--80, 2014.

\bibitem[Chunxiang et~al.(2016)Chunxiang, Li, and Wang]{chunxiang2016optimal}
A~Chunxiang, Zhongfei Li, and Fan Wang.
\newblock Optimal investment strategy under time-inconsistent preferences and
  high-water mark contract.
\newblock \emph{Operations Research Letters}, 44\penalty0 (2):\penalty0
  212--218, 2016.

\bibitem[Chen et~al.(2018)Chen, Li, and Zeng]{chen2018optimal}
Shumin Chen, Zhongfei Li, and Yan Zeng.
\newblock Optimal dividend strategy for a general diffusion process with
  time-inconsistent preferences and ruin penalty.
\newblock \emph{Journal on Financial Mathematics}, 9\penalty0 (1):\penalty0
  274--314, 2018.

\bibitem[van Hasselt et~al.(2019)van Hasselt, Quan, Hessel, Xu, Borsa, and
  Barreto]{van2019general}
Hado van Hasselt, John Quan, Matteo Hessel, Zhongwen Xu, Diana Borsa, and
  Andr{\'e} Barreto.
\newblock General non-linear {B}ellman equations.
\newblock \emph{arXiv preprint arXiv:1907.03687}, 2019.

\bibitem[Ng et~al.(2000)Ng, Russell, et~al.]{ng2000algorithms}
Andrew~Y Ng, Stuart~J Russell, et~al.
\newblock Algorithms for inverse reinforcement learning.
\newblock In \emph{{International Conference on Machine Learning}}, volume~1,
  2000.

\bibitem[Abbeel and Ng(2004)]{abbeel2004apprenticeship}
Pieter Abbeel and Andrew~Y Ng.
\newblock Apprenticeship learning via inverse reinforcement learning.
\newblock In \emph{{International Conference on Machine Learning}}, volume~5,
  2004.

\bibitem[Ziebart et~al.(2008)Ziebart, Maas, Bagnell, Dey,
  et~al.]{ziebart2008maximum}
Brian~D Ziebart, Andrew~L Maas, J~Andrew Bagnell, Anind~K Dey, et~al.
\newblock Maximum entropy inverse reinforcement learning.
\newblock In \emph{{AAAI}}, volume~23, pages 1433--1438, 2008.

\bibitem[Garg et~al.(2021)Garg, Chakraborty, Cundy, Song, and
  Ermon]{garg2021iq}
Divyansh Garg, Shuvam Chakraborty, Chris Cundy, Jiaming Song, and Stefano
  Ermon.
\newblock {IQ-Learn}: {I}nverse soft-q learning for imitation.
\newblock In \emph{{Advances in Neural Information Processing Systems}},
  volume~34, pages 4028--4039. Curran Associates, Inc., 2021.

\bibitem[Mombaur et~al.(2010)Mombaur, Truong, and Laumond]{mombaur2010human}
Katja Mombaur, Anh Truong, and Jean-Paul Laumond.
\newblock From human to humanoid locomotion -- an inverse optimal control
  approach.
\newblock \emph{{Autonomous Robots}}, 28\penalty0 (3):\penalty0 369--383, 2010.

\bibitem[Rothkopf and Ballard(2013)]{rothkopf2013modular}
Constantin~A Rothkopf and Dana~H Ballard.
\newblock Modular inverse reinforcement learning for visuomotor behavior.
\newblock \emph{Biological cybernetics}, 107\penalty0 (4):\penalty0 477--490,
  2013.

\bibitem[Muelling et~al.(2014)Muelling, Boularias, Mohler, Sch{\"o}lkopf, and
  Peters]{muelling2014learning}
Katharina Muelling, Abdeslam Boularias, Betty Mohler, Bernhard Sch{\"o}lkopf,
  and Jan Peters.
\newblock Learning strategies in table tennis using inverse reinforcement
  learning.
\newblock \emph{{Biological Cybernetics}}, 108\penalty0 (5):\penalty0 603--619,
  2014.

\bibitem[Schultheis et~al.(2021)Schultheis, Straub, and
  Rothkopf]{schultheis2021inverse}
Matthias Schultheis, Dominik Straub, and Constantin~A Rothkopf.
\newblock Inverse optimal control adapted to the noise characteristics of the
  human sensorimotor system.
\newblock In \emph{{Advances in Neural Information Processing Systems}},
  volume~34, pages 9429--9442. Curran Associates, Inc., 2021.

\bibitem[Golub et~al.(2013)Golub, Chase, and Yu]{golub2013learning}
Matthew Golub, Steven Chase, and Byron Yu.
\newblock Learning an internal dynamics model from control demonstration.
\newblock In \emph{{International Conference on Machine Learning}}, pages
  606--614. PMLR, 2013.

\bibitem[Ashwood et~al.(2020)Ashwood, Roy, Bak, and
  Pillow]{ashwood2020inferring}
Zoe Ashwood, Nicholas~A. Roy, Ji~Hyun Bak, and Jonathan~W Pillow.
\newblock Inferring learning rules from animal decision-making.
\newblock In \emph{{Advances in Neural Information Processing Systems}},
  volume~33, pages 3442--3453. Curran Associates, Inc., 2020.

\bibitem[Aalen et~al.(2008)Aalen, Borgan, and Gjessing]{aalen2008survival}
Odd Aalen, Ornulf Borgan, and Hakon Gjessing.
\newblock \emph{Survival and event history analysis: a process point of view}.
\newblock {Springer Science \& Business Media}, 2008.

\bibitem[Hanson(2007)]{hanson2007applied}
Floyd~B Hanson.
\newblock \emph{Applied stochastic processes and control for jump-diffusions:
  modeling, analysis and computation}.
\newblock SIAM, 2007.

\bibitem[S{\"a}rkk{\"a} and Solin(2019)]{sarkka2019applied}
Simo S{\"a}rkk{\"a} and Arno Solin.
\newblock \emph{Applied stochastic differential equations}, volume~10.
\newblock Cambridge University Press, 2019.

\bibitem[Paszke et~al.(2019)Paszke, Gross, Massa, Lerer, Bradbury, Chanan,
  Killeen, Lin, Gimelshein, Antiga, Desmaison, Kopf, Yang, DeVito, Raison,
  Tejani, Chilamkurthy, Steiner, Fang, Bai, and Chintala]{paszke2019pytorch}
Adam Paszke, Sam Gross, Francisco Massa, Adam Lerer, James Bradbury, Gregory
  Chanan, Trevor Killeen, Zeming Lin, Natalia Gimelshein, Luca Antiga, Alban
  Desmaison, Andreas Kopf, Edward Yang, Zachary DeVito, Martin Raison, Alykhan
  Tejani, Sasank Chilamkurthy, Benoit Steiner, Lu~Fang, Junjie Bai, and Soumith
  Chintala.
\newblock Pytorch: {A}n imperative style, high-performance deep learning
  library.
\newblock In \emph{{Advances in Neural Information Processing Systems}},
  volume~32. Curran Associates, Inc., 2019.

\bibitem[Rabitz et~al.(1983)Rabitz, Kramer, and Dacol]{rabitz1983sensitivity}
Herschel Rabitz, Mark Kramer, and D~Dacol.
\newblock Sensitivity analysis in chemical kinetics.
\newblock \emph{{Annual Review of Physical Chemistry}}, 34\penalty0
  (1):\penalty0 419--461, 1983.

\bibitem[Kim et~al.(2021)Kim, Shin, and Yang]{kim2021hamilton}
Jeongho Kim, Jaeuk Shin, and Insoon Yang.
\newblock Hamilton-{J}acobi deep {Q}-learning for deterministic continuous-time
  systems with {L}ipschitz continuous controls.
\newblock \emph{Journal of Machine Learning Research}, 22:\penalty0 206--1,
  2021.

\bibitem[Merton(1976)]{merton1976option}
Robert~C Merton.
\newblock Option pricing when underlying stock returns are discontinuous.
\newblock \emph{{Journal of Financial Economics}}, 3\penalty0 (1-2):\penalty0
  125--144, 1976.

\bibitem[K~Namboodiri et~al.(2014)K~Namboodiri, Mihalas, Marton, and
  Hussain~Shuler]{k2014general}
Vijay~Mohan K~Namboodiri, Stefan Mihalas, Tanya Marton, and Marshall~Gilmer
  Hussain~Shuler.
\newblock A general theory of intertemporal decision-making and the perception
  of time.
\newblock \emph{{Frontiers in Behavioral Neuroscience}}, 8:\penalty0 61, 2014.

\end{thebibliography}
